\documentclass{article}
\pdfoutput=1
\usepackage[margin=1.0in]{geometry}
\usepackage{graphicx}
\usepackage{tikz}
\usepackage{comment}
\usepackage{amsmath,amssymb} 
\usepackage{color}

\usepackage[numbers,sort&compress]{natbib}

\usepackage{bm}
\usepackage{arydshln}
\usepackage{booktabs}
\usepackage{physics}
\usepackage{multirow}
\usepackage{amsmath}
\DeclareMathOperator*{\argmax}{argmax}
\makeatletter
\newcommand*\bigcdot{\mathpalette\bigcdot@{.7}}
\newcommand*\bigcdot@[2]{\mathbin{\vcenter{\hbox{\scalebox{#2}{$\m@th#1\bullet$}}}}}
\makeatother

\usepackage[linesnumbered,ruled,vlined]{algorithm2e}

\SetCommentSty{mycommfont}
\usepackage{caption}
\usepackage{subcaption}

\SetKwInput{KwInput}{Input}                %
\SetKwInput{KwOutput}{Output}              %
\usepackage{float}
\usepackage{url}

\title{Adversarial amplitude swap towards robust image classifiers} %

\author {
    Chun Yang Tan,\footnote{Graduate School of Science and Engineering, Chiba University (email: chunyangtan@gmail.com)} 
    \ Kazuhiko Kawamoto$^{\dagger}$
    \ Hiroshi Kera,\footnote{Graduate School of Engineering, Chiba University (email: kawa@faculty.chiba-u.jp, kera.hiroshi@gmail.com)} 
}
\date{}

\begin{document}

\maketitle

\begin{abstract}
The vulnerability of convolutional neural networks (CNNs) to image perturbations such as common corruptions and adversarial perturbations has recently been investigated from the perspective of frequency. In this study, we investigate the effect of the amplitude and phase spectra of adversarial images on the robustness of CNN classifiers. Extensive experiments revealed that the images generated by combining the amplitude spectrum of adversarial images and the phase spectrum of clean images accommodates moderate and general perturbations, and training with these images equips a CNN classifier with more general robustness, performing well under both common corruptions and adversarial perturbations. We also found that two types of overfitting (catastrophic overfitting and robust overfitting) can be circumvented by the aforementioned spectrum recombination. We believe that these results contribute to the understanding and the training of truly robust classifiers.

\end{abstract}

\section{Introduction}
\begin{figure}[t]
    \centering
    \includegraphics[width=0.9\textwidth]{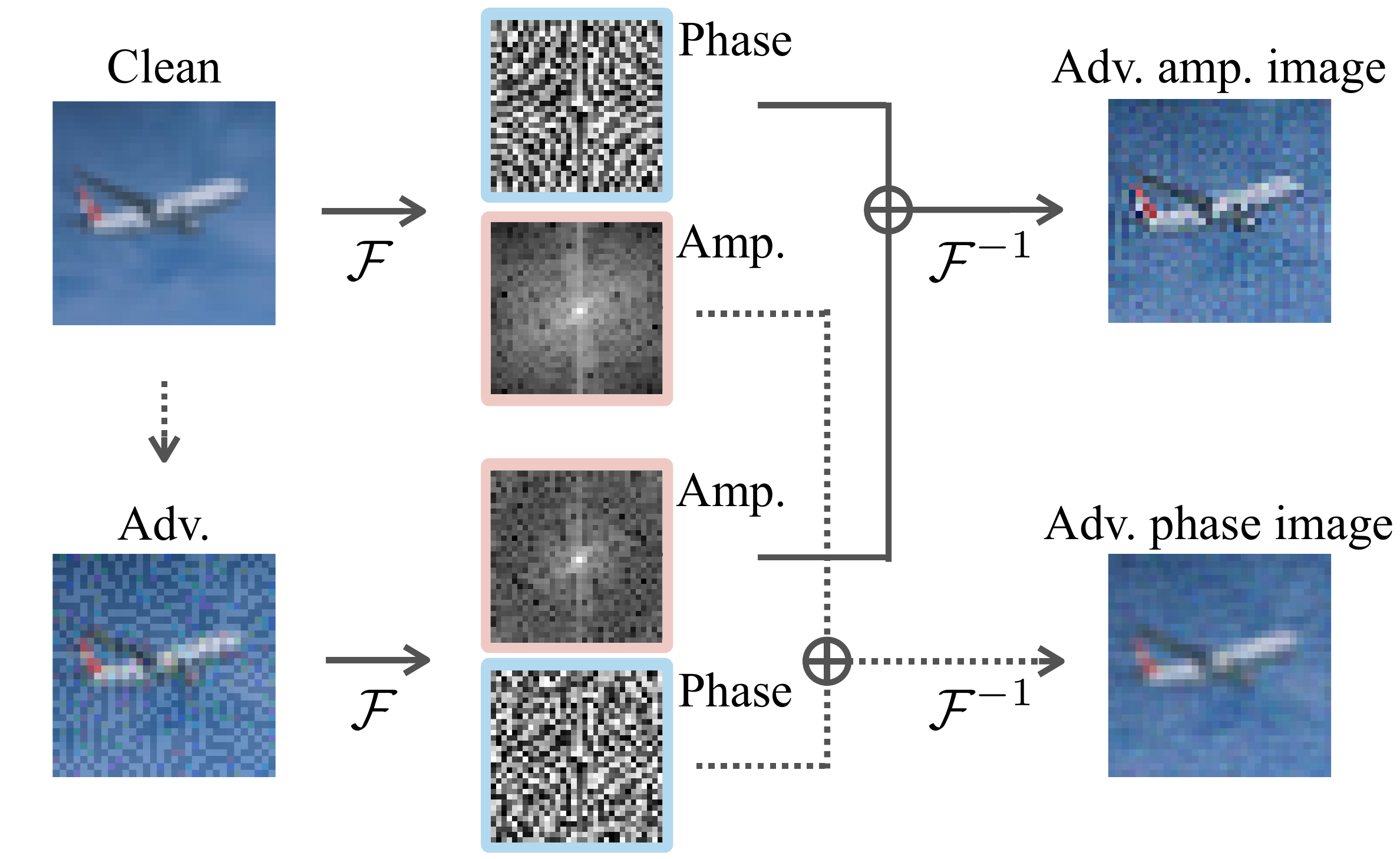}
\caption{The pipeline of the proposed frequency-based data augmentation. The maps $\mathcal{F}$ and $\mathcal{F}^{-1}$ denote the discrete Fourier transform and its inverse. The amplitude spectra of a clean image and its adversarial image are swapped to generate adversarial amplitude and adversarial phase images. 
}\label{fig:teaser}
\end{figure}
Despite their state-of-the-art performance~\cite{krizhevsky2012imagenet,lecun1998gradient}, convolutional neural networks~(CNNs) have been found to be vulnerable to perturbations in images such as common corruptions~\cite{vasiljevic2016examining,geirhos2018generalisation,hendrycks2019corruption}~(e.g., fog corruption) and adversarial perturbations~\cite{szegedy2014intriguing,carlini2017towards,carlini2017adversarial,kurakin2018adversarialphysical,narodytska2017blackbox,papernot2017practicalblackbox}. Although such perturbations do not change the semantic information of images, they substantially degrade the performance of CNN classifiers. 
Several recent studies attempted to explain and resolve this issue from the frequency perspective. 
For example, \citet{chen2021apr} revealed that CNN classifiers rely more on the amplitude spectrum of images rather than the phase spectrum, whereas humans rely more on the phase spectrum. Based on this observation, they proposed amplitude-phase-recombination~(APR), a data augmentation method that randomly replaces the amplitude spectrum of images with that of others in the training set. The APR encourages CNNs to focus more on the phase information, as humans do, leading to higher robustness of CNNs against common corruptions. 
It has also been shown that CNNs can be trained to perform better in several image classification and object recognition tasks by swapping or altering a part of the amplitude spectrum of images~\cite{xu2021domaingeneralization, yang2020fourierda}.

The vulnerability of CNNs to adversarial perturbations has also been discussed from the frequency perspective. \citet{wang2020highfrequency} suggested that CNNs exploit high-frequency image components that are not perceptible to humans. However, more recent studies claimed that adversarial perturbations are neither high nor low-frequency components, but rather are dataset dependent~\cite{maiya2021frequency, tsuzuku2019fourierbasis}. These studies focused on the relationships between frequency bands and adversarial perturbations, and the effect of the amplitude and phase spectra of adversarial images on the robustness of CNN classifiers remains unexplored.

In this study, we investigate the nature of the amplitude and phase spectra of adversarial images. Inspired by~\cite{chen2021apr}, we propose a frequency-based data augmentation method in which the amplitude spectra are swapped between clean and adversarial images to generate adversarial amplitude and adversarial phase images~(Fig.~\ref{fig:teaser}). In particular, training with the former images is expected to provide more general robustness because it encourages a CNN classifier to focus on the phase spectrum, which contains semantic information, and also resist the adversarially-designed amplitude spectrum. 
We verified this idea through extensive experiments, and revealed that the training with the adversarial amplitude images achieved the followings: 
\begin{itemize}
\item[$\bigcdot$] It led to CNN classifiers that are robust to both common corruptions and adversarial perturbations, whereas the baseline methods only enhance either of them. 
\item[$\bigcdot$] It also helped the CNN classifiers learn from moderate, strong, and even extremely strong adversarial images, whereas both the standard training with adversarial images (i.e., adversarial training) and that with adversarial phase images suffer from catastrophic overfitting~\cite{tramer2018ensembleattdef,kim2020understandingco, kang2021understandingco}, leading to poor performance on adversarial perturbations other than those specifically trained against.  
\item[$\bigcdot$] It prevented the catastrophic and the robust overfitting during the adversarial training, whereas each of the other convention data augmentation methods, the random crop, and horizontal flip, did not. 
\end{itemize}

The experimental results show that (i) the amplitude spectrum of adversarial images accommodates moderate and general perturbations that helped classifiers to equip more general robustness, and (ii) the phase spectrum of adversarial images tended to be moderate but still adversarial, which may improve the adversarial robustness of classifiers but also retain the risk of catastrophic overfitting.
We believe that this study deepens the understanding of general perturbations in images, particularly from the frequency perspective, and contributes to the future development of truly robust image classifiers. 

\section{Related Work}
\paragraph{The utility of amplitude and phase spectra.} Earlier studies empirically showed that the semantic information required for humans to recognize objects is contained more in the phase spectrum of images than in the amplitude spectrum~\cite{phaseimpor,phasesaliency}.
However, it was very recently shown by \citet{chen2021apr} that this is not the case with CNN classifiers. They generated a fused image that consisted of the phase spectrum of an image of a \textit{Revolver} and the amplitude spectrum of an image of a \textit{Jigsaw Puzzle}. Notably, the CNN classifier predicted the label of the fused image as \textit{Jigsaw Puzzle}. This indicates that, unlike humans, the prediction of CNN models is generally determined by the amplitude spectrum rather than the phase spectrum. 
This disparity between humans and CNN classifiers has motivated several studies to examine the influence of amplitude and phase spectra during the training of CNNs. \citet{chen2021apr} proposed the APR data augmentation to enhance the robustness of CNN classifiers, where the phase spectrum of an image and the amplitude spectrum of another image in the training dataset are recombined to generate a new image, the label of which is set to that of the original image. 
It was also shown that by swapping or altering the amplitude spectrum of training images, CNNs can achieve better performance in semantic segmentation and domain generalization tasks~\cite{xu2021domaingeneralization,yang2020fourierda}. These studies perturbed the amplitude spectrum of the original images in various ways while keeping the phase spectrum unchanged. This strategy encourages CNNs to learn semantic information from the phase spectrum, and become stable to changes in the amplitude spectrum. 

\paragraph{Adversarial images and the frequency bands.} \citet{ilyas2019notbugbutfeatures} first suggested that adversarial vulnerability is caused by the perceptual disparity between humans and CNNs; namely, CNNs view the images at a much higher granularity than a human. \citet{wang2020highfrequency} then provided a more concrete explanation of this disparity; that is, CNNs exploit the high-frequency image components that are not perceptible to humans. This led to the speculation that adversarial perturbations are highly related to high-frequency components. However, \citet{tsuzuku2019fourierbasis} and \citet{maiya2021frequency} showed that most CNNs are sensitive to Fourier basis perturbations regardless of high or low-frequency components. Model robustness against Fourier basis perturbations was then explained and visualized by \citet{yin2019modelrobustness}. Moreover, \citet{guo2020lowfreqperturbation} showed that adversarial attacks can still be effective even when the adversarial subspace is restricted to the low end of the frequency spectrum. 
Although these studies discussed adversarial images in terms of frequency, to the best of our knowledge, the nature of the amplitude and phase spectra of adversarial images remains unexplored. In this study, we investigate these aspects, particularly in terms of their effect on the general robustness or generalization ability of CNN classifiers.

\section{Preliminaries}
\paragraph{Amplitude and phase spectrum.}
Let $\bm{x}$ be an $N\times N$ image. The discrete Fourier transform (DFT) and its inverse~(iDFT) are defined as follows~\cite{bracewell1986fouriertransform}. 
\begin{align}
    \mathcal{F}(u,v)&=\sum^{N-1}_{h=0}\sum^{N-1}_{w=0} \bm{x}(h,w)e^{-i2\pi\left(uh +vw\right)/N},\\ 
    x(h,w) &= \frac{1}{N\times N}\sum^{N-1}_{u=0}\sum^{N-1}_{v=0} \mathcal{F}(u,v) e^{i2\pi\left(uh +vw\right)/N},
\end{align}
where $\bm{x}(h,w)$ denotes the $(h,w)$-th pixel value of $\bm{x}$, and $F(u, v)$ denotes the $(u, v)$-th DFT component of $\bm{x}$. Both the DFT and iDFT are computed with the fast Fourier transform (FFT) algorithm~\cite{fft}, which is provided in the PyTorch library~\cite{paszke2019pytorch}.
The amplitude spectrum $\mathcal{A}(\bm{x})$ and phase spectrum $\mathcal{P}(\bm{x})$ are then represented as follows.
\begin{align}
    \mathcal{A}(\bm{x}) &= \sqrt {\text{Re}[\mathcal{F}(\bm{x})]^2 + \text{Im}[\mathcal{F}(\bm{x})]^2},\label{eq:amp}
    \\
    \mathcal{P}(\bm{x}) &= \arctan \left( \frac{\text{Im}[\mathcal{F}(\bm{x})]}{\text{Re}[\mathcal{F}(\bm{x})]} \right),\label{eq:phase}
\end{align}
where $\text{Re}[\mathcal{F}(\bm{x})]$ and $\text{Im}[\mathcal{F}(\bm{x})]$ denote the real and imaginary parts of $\mathcal{F}(\bm{x})$, respectively. For RGB images, the DFT is performed for each channel to obtain the corresponding amplitude and phase spectrum. Hereinafter, we denote $\mathcal{A}(\bm{x})$ and $\mathcal{P}(\bm{x})$ by the amplitude spectrum and the phase spectrum of image $\bm{x}$, respectively. 

\paragraph{Adversarial images.} Let $\mathcal{X}$ be the image domain, and let $\mathcal{Y}=\{1, 2, \ldots, K\}$. Let $F(\bm{x}) = \argmax_{i\in\mathcal{Y}} f_i(\bm{x})$ be a $K$-class CNN classifier, where $f_i:\mathcal{X}\to\mathbb{R}$ denotes the $i$-th logit of the CNN. Given a distance function $d:\mathcal{X}\times \mathcal{X}\to \mathbb{R}_{\ge 0}$ and a budget $\epsilon > 0$, an adversarial image $\bm{x}_{\text{adv}}\in\mathcal{X}$ of $\bm{x}\in\mathcal{X}$ with the label $y\in\mathcal{Y}$ is an image such that $F(\bm{x}_{\text{adv}}) \ne y$ and $d(\bm{x}, \bm{x}_{\text{adv}}) \le \epsilon$.
Gradient-based non-targeted methods to generate adversarial examples typically solve the following problem approximately based on gradient ascent with some projections. 
\begin{align}\label{eq:grad-based-adv}
    \max_{\bm{x}'\in\mathcal{X}}\ \ell(f(\bm{x}), y),\quad\text{s.t.}\  d(\bm{x},\bm{x}') \le \epsilon,
\end{align}
where $f(\bm{x}) = (f_1(\bm{x}), \ldots, f_K(\bm{x}))^{\top}$, and $\ell(\,\cdot\,,\,\cdot\,)$ denotes a loss function (e.g., cross-entropy loss). 
In this study, we use two standard methods, the fast gradient sign method~(FGSM~\cite{goodfellow2015explaining}) and the projected gradient descent~(PGD~\cite{madry2018towards}). For the PGD methods, we use two typical metrics, $l_2$ and $l_\infty$ norms to measure the magnitude of perturbations and denote them by PGD-$l_2$ and PGD-$l_\infty$ respectively.

\paragraph{Adversarial training and overfitting.}
Adversarial training~\cite{szegedy2014intriguing} is a defense method in which a classifier is trained to classify adversarial images correctly. In particular, adversarial training with FGSM is a computationally inexpensive approach, and thus, there have been various studies for improvements~\cite{wong2020fast,andriushchenko2020improvefast}. To resist strong adversarial images, it is necessary to use an FGSM with a large $\epsilon$ in training. However, it is known that when $\epsilon$ is too large, the classifier starts performing unreasonably well for FGSM adversarial images and fails to classify clean or other adversarial images (e.g., those generated by PGD). This is known as catastrophic overfitting~\cite{tramer2018ensembleattdef,kim2020understandingco, kang2021understandingco}. Another form of overfitting in adversarial training, robust overfitting, was also recently pointed out, in which the classification accuracy of adversarial images starts degrading substantially at some point in an adversarial training with the PGD~\cite{madry2018towards}. 
The development of methods to circumvent these forms of overfitting and train a strongly defended classifier remains an unresolved challenge. 

\section{Adversarial Amplitude Swap}
To train a CNN classifier to be robust against both common corruptions and adversarial perturbations, we consider that both the amplitude and phase spectra of images play important roles. This idea motivated a new frequency-based data augmentation method, called the adversarial amplitude swap. Given a clean and its adversarial image, this method swaps the amplitude spectrum of the former with that of the latter to generate two augmented images: an adversarial amplitude image, which has the amplitude of the adversarial image and the phase of the clean image, and an adversarial phase image, which is the opposite~(Fig.~\ref{fig:teaser}). 

Formally, the process of the adversarial amplitude swap is performed as follows. 
First,  given a clean image $\bm{x}$, an adversarial image $\bm{x}_{\text{adv}}$ is generated. Then, the DFT is applied to the two images to obtain the amplitude--phase decompositions, $(\mathcal{A}(\bm{x}), \mathcal{P}(\bm{x}))$ and $(\mathcal{A}(\bm{x}_{\text{adv}}), \mathcal{P}(\bm{x}_{\text{adv}}))$.
The adversarial amplitude and adversarial phase images, $x_{\text{AA}}$ and $x_{\text{AP}}$, are then constructed by the inverse DFT of $(\mathcal{A}(\bm{x}), \mathcal{P}(\bm{x}_{\text{adv}}))$ and $(\mathcal{A}(\bm{x}_{\text{adv}}), \mathcal{P}(\bm{x}))$, respectively; namely, 
\begin{align}
    \bm{x}_{\text{AA}} &= \mathcal{F}^{-1}\qty(\mathcal{A}(\bm{x}_{\text{adv}})\cdot e^{i\cdot \mathcal{P}(\bm{x})}), \label{eq:aa}\\
    \bm{x}_{\text{AP}} &= \mathcal{F}^{-1}\qty(\mathcal{A}(\bm{x})\cdot e^{i\cdot \mathcal{P}(\bm{x}_{\text{adv}})}).\label{eq:ap}
\end{align}
Pseudo-code for this process is provided in Algorithm~\ref{alg:AAS}. 

Note that the adversarial image of the clean image changes at each training step because its generation depends on the classifier (cf. Eq.~\eqref{eq:grad-based-adv}). Thus, $\bm{x}_{\text{AA}}$ has a static phase spectrum derived from $\bm{x}$, and a stochastic amplitude spectrum derived from different $\bm{x}_{\text{adv}}$, along the training. Therefore, training with $\bm{x}_{\text{AA}}$ encourages CNN classifiers to learn the static semantic features from the phase spectrum and also resist the stochastic adversarial features in the amplitude spectrum.
Similarly, $\bm{x}_{\text{AP}}$ has a static amplitude spectrum of a clean image and the stochastic phase spectrum of the adversarial image, which encourages the classifiers to learn more on the amplitude spectrum and resist the adversarial features in the phase spectrum. In Sections~\ref{sec:experiment-classification} and~\ref{sec:experiment-ablation}, we show that training with adversarial amplitude images, $\bm{x}_{\text{AA}}$, leads to robust CNN classifiers against both common corruptions and adversarial perturbations. 
Another advantage of the use of the adversarial amplitude images is that it stabilizes the training to circumvent catastrophic and robust overfitting for moderate, strong, and even extremely strong levels of attack strength~(cf. Sections~\ref{sec:experiment-extreme} and~\ref{sec:experiment-overfitting}). 

\begin{algorithm}[t]
\DontPrintSemicolon
  \KwInput{$\bm{x}$: clean image}
  \KwOutput{$\bm{x}_{\text{AA}}$: adversarial amplitude image, $\bm{x}_{\text{AP}}$: adversarial phase image}
  $\bm{x}_{\text{adv}} \leftarrow \textsc{AdversarialAttack}(\bm{x})$. \tcp*{E.g., by the FGSM.}  
  $\mathcal{A}(\bm{x}), \mathcal{P}(\bm{x}) \leftarrow \textsc{DFT}(\bm{x})$  \tcp*{By Eqs.~\eqref{eq:amp} and~\eqref{eq:phase}}  
  $\mathcal{A}(\bm{x}_{\text{adv}}), \mathcal{P}(\bm{x}_{\text{adv}}) \leftarrow \textsc{DFT}(\bm{x}_{\text{adv}})$ \\ 
  $\bm{x}_{\text{AA}} \leftarrow \textsc{InvDFT}(\mathcal{A}(\bm{x}_{\text{adv}}), \mathcal{P}(\bm{x}))$ \tcp*{By Eqs.~\eqref{eq:aa} and~\eqref{eq:ap}} 
  $\bm{x}_{\text{AP}} \leftarrow \textsc{InvDFT}(\mathcal{A}(\bm{x}), \mathcal{P}(\bm{x}_{\text{adv}}))$ 
\caption{Adversarial amplitude swap}\label{alg:AAS}
\end{algorithm}

\section{Experiments}
\subsection{Experiment Setup}
\paragraph{Datasets.} 
We conducted multiple experiments on CIFAR-10 and CIFAR-100 datasets~\cite{krizhevsky2009multilayersfeatures}. 
To measure the robustness against common corruptions, we evaluated methods on the CIFAR-10-C and CIFAR-100-C datasets~\cite{hendrycks2019corruption}. The CIFAR-10-C and CIFAR-100-C datasets were constructed by corrupting the original CIFAR-10 and CIFAR-100 test sets, respectively. A total of 15 types of noise are included, such as \textit{gaussian noise, shot noise, impulse noise, defocus blur, frosted, glass blur, motion blur, zoom blur, snow, frost, fog, brightness, contrast, elastic, pixelate, JPEG}, each appearing at five severity levels or intensities. Level-1 represents the lowest severity and level-5 represents the highest severity. For evaluation of adversarial robustness, FGSM attacks with $\epsilon=\epsilon_0/255$ for $\epsilon_0 \in \{4,8,16,32\}$, PGD-$l_\infty$ attack with $\epsilon=8/255, \text{ the step size }\alpha=0.1, \text{ and the number of iterations }i_\text{iters}=20$, and PGD-$l_2$ attack with $\epsilon=0.5, \alpha=0.1, i_\text{iters}=20$ were used.

\paragraph{Training setup.}
We trained classifiers with several network architectures, including ResNet-18~\cite{he2016resnet}, WideResNet-40-2~\cite{zagoruyko2016wrn} and DenseNet~\cite{huang2017densenet}. The learning rate was set to 0.01 with a decay of 0.1 at the 100th and 150th epochs. The classifiers were optimized with the stochastic gradient descent using a momentum of 0.9, and a weight decay of $5\times 10^{-4}$. Two data augmentation methods, including random crop and random horizontal flip, were implemented. For the generation of adversarial robustness, we used Foolbox~\cite{rauber2017foolbox}.

\paragraph{Baselines.}  
We adopted two data-augmentation-based methods as the baselines.  The first baseline method is the APR-SP~\cite{chen2021apr} (simply denoted by APR), which is a frequency-based data augmentation method for enhancing the classifier robustness against common corruptions.\footnote{A small experiment on the adversarial training with the APR-SP was also conducted in~\cite{chen2021apr}. However, the results were not as good as other baseline methods in our experiments. Due to the page limitations, we send it to the supplementary material.} The official implementation was used.\footnote{\url{https://github.com/iCGY96/APR}} 
The second baseline method is the standard training with clean and adversarial images, which corresponds to GoodFellow's adversarial training with the weight of $\alpha=0.5$~\cite{goodfellow2015explaining}, which aims to enhance the classifier robustness against adversarial perturbations. 

\subsection{CIFAR-10 \& CIFAR-100 Image Classification}\label{sec:experiment-classification}
\setlength{\tabcolsep}{7pt}
\begin{table}[p]
    \centering
    \caption{The classification accuracy (\%) of ResNet-18, WideResNet-40-2, and DenseNet classifiers trained on CIFAR-10 with different combination of images. The FGSM attack with $\epsilon=8/255$ was used in training. The top-2 results are indicated in bold while the best results are underlined.}
    \label{table:cifar10_fgsm8}
    \begin{tabular}{ccccccc}
        \hline
        & \multicolumn{1}{c|}{}& \multicolumn{5}{c}{Combination of Training Data} \\
        &\multicolumn{1}{c|}{}&Clean&APR&\multicolumn{1}{c|}{C\&Adv}&C\&AA&C\&AP\\
        \hline
        \multicolumn{1}{c|}{\multirow{11}{*}{ResNet-18}}&\multicolumn{1}{c|}{Clean} &   
            \underline{\textbf{94.8}}&\textbf{94.6}&\multicolumn{1}{c|}{93.3}&92.8&89.1\\
        \cline{2-7}
        \multicolumn{1}{c|}{}&\multicolumn{1}{c|}{FGSM ($\epsilon_0=4$)} &
            71.6&70.9&\multicolumn{1}{c|}{\underline{\textbf{90.8}}}&\textbf{82.3}&74.8\\
        \multicolumn{1}{c|}{}&\multicolumn{1}{c|}{FGSM ($\epsilon_0=8$)} &
            65.6&66.2&\multicolumn{1}{c|}{\underline{\textbf{87.4}}}&\textbf{78.5}&69.8\\
        \multicolumn{1}{c|}{}&\multicolumn{1}{c|}{FGSM ($\epsilon_0=16$)} &
            58.1&61.3&\multicolumn{1}{c|}{\underline{\textbf{80.3}}}&\textbf{74.0}&63.7\\
        \multicolumn{1}{c|}{}&\multicolumn{1}{c|}{FGSM ($\epsilon_0=32$)} &
            44.7&54.3&\multicolumn{1}{c|}{\underline{\textbf{72.4}}}&\textbf{68.2}&57.0\\
        \multicolumn{1}{c|}{}&\multicolumn{1}{c|}{PGD-$l_\infty$} &
            0.0&0.3&\multicolumn{1}{c|}{3.4}&\textbf{30.6}&\underline{\textbf{33.5}}\\
        \multicolumn{1}{c|}{}&\multicolumn{1}{c|}{PGD-$l_2$} &
            1.1&7.7&\multicolumn{1}{c|}{7.9}&\textbf{54.3}&\underline{\textbf{57.0}}\\
        \cline{2-7}
        \multicolumn{1}{c|}{}&\multicolumn{1}{c|}{Corrupted-1} &   
            \textbf{91.2}&\underline{\textbf{92.7}}&\multicolumn{1}{c|}{\textbf{91.2}}&\textbf{91.2}&88.9\\
        \multicolumn{1}{c|}{}&\multicolumn{1}{c|}{Corrupted-2} &   
            89.0&\underline{\textbf{91.4}}&\multicolumn{1}{c|}{89.7}&\textbf{90.0}&87.8\\
        \multicolumn{1}{c|}{}&\multicolumn{1}{c|}{Corrupted-3} &   
            87.1&\underline{\textbf{90.3}}&\multicolumn{1}{c|}{88.5}&\textbf{89.0}&86.7\\
        \multicolumn{1}{c|}{}&\multicolumn{1}{c|}{Corrupted-4} &   
            84.5&\underline{\textbf{88.5}}&\multicolumn{1}{c|}{86.5}&\textbf{87.4}&85.4\\
        \multicolumn{1}{c|}{}&\multicolumn{1}{c|}{Corrupted-5} &   
            80.9&\underline{\textbf{85.9}}&\multicolumn{1}{c|}{83.7}&\textbf{84.9}&82.9\\
        \hline 
        \hline
        \multicolumn{1}{c|}{\multirow{11}{*}{WideResNet}}&\multicolumn{1}{c|}{Clean} &   
            \textbf{94.1}&\underline{\textbf{94.3}}&\multicolumn{1}{c|}{86.1}&91.3&88.3\\
        \cline{2-7}
        \multicolumn{1}{c|}{}&\multicolumn{1}{c|}{FGSM ($\epsilon_0=4$)} &
            \textbf{72.5}&69.7&\multicolumn{1}{c|}{68.5}&\underline{\textbf{76.5}}&70.9\\
        \multicolumn{1}{c|}{}&\multicolumn{1}{c|}{FGSM ($\epsilon_0=8$)} &
            \textbf{66.8}&64.5&\multicolumn{1}{c|}{63.7}&\underline{\textbf{72.4}}&65.7\\
        \multicolumn{1}{c|}{}&\multicolumn{1}{c|}{FGSM ($\epsilon_0=16$)} &
            58.8&60.0&\multicolumn{1}{c|}{59.0}&\underline{\textbf{67.0}}&\textbf{60.3}\\
        \multicolumn{1}{c|}{}&\multicolumn{1}{c|}{FGSM ($\epsilon_0=32$)} &
            43.5&51.0&\multicolumn{1}{c|}{53.5}&\underline{\textbf{60.3}}&\textbf{54.3}\\
        \multicolumn{1}{c|}{}&\multicolumn{1}{c|}{PGD-$l_\infty$} &
            0.2&0.3&\multicolumn{1}{c|}{\underline{\textbf{45.7}}}&28.7&\textbf{38.4}\\
        \multicolumn{1}{c|}{}&\multicolumn{1}{c|}{PGD-$l_2$} &
            2.4&7.0&\multicolumn{1}{c|}{\textbf{53.8}}&51.7&\underline{\textbf{54.2}}\\
        \cline{2-7}
        \multicolumn{1}{c|}{}&\multicolumn{1}{c|}{Corrupted-1} &   
        \textbf{90.4}&\underline{\textbf{92.2}}&\multicolumn{1}{c|}{86.2}&89.4&87.9\\
        \multicolumn{1}{c|}{}&\multicolumn{1}{c|}{Corrupted-2} &   
        88.1&\underline{\textbf{90.9}}&\multicolumn{1}{c|}{85.0}&\textbf{88.2}&86.7\\
        \multicolumn{1}{c|}{}&\multicolumn{1}{c|}{Corrupted-3} &   
        85.9&\underline{\textbf{89.7}}&\multicolumn{1}{c|}{83.8}&\textbf{86.9}&85.4\\
        \multicolumn{1}{c|}{}&\multicolumn{1}{c|}{Corrupted-4} &   
        83.2&\underline{\textbf{87.7}}&\multicolumn{1}{c|}{82.2}&\textbf{85.3}&83.8\\
        \multicolumn{1}{c|}{}&\multicolumn{1}{c|}{Corrupted-5} &   
        79.3&\underline{\textbf{85.1}}&\multicolumn{1}{c|}{79.5}&\textbf{82.4}&81.2\\
        \hline
        \hline
        \multicolumn{1}{c|}{\multirow{11}{*}{DenseNet}}&\multicolumn{1}{c|}{Clean} &   
            \textbf{93.4}&\underline{\textbf{93.7}}&\multicolumn{1}{c|}{87.4}&91.9&88.8\\
        \cline{2-7}
        \multicolumn{1}{c|}{}&\multicolumn{1}{c|}{FGSM ($\epsilon_0=4$)} &
            62.6&64.18&\multicolumn{1}{c|}{\textbf{64.7}}&\underline{\textbf{75.7}}&64.3\\
        \multicolumn{1}{c|}{}&\multicolumn{1}{c|}{FGSM ($\epsilon_0=8$)} &
            57.1&\textbf{59.2}&\multicolumn{1}{c|}{56.6}&\underline{\textbf{70.6}}&56.7\\
        \multicolumn{1}{c|}{}&\multicolumn{1}{c|}{FGSM ($\epsilon_0=16$)} &
            50.2&\textbf{55.6}&\multicolumn{1}{c|}{49.9}&\underline{\textbf{64.9}}&50.1\\
        \multicolumn{1}{c|}{}&\multicolumn{1}{c|}{FGSM ($\epsilon_0=32$)} &
            39.1&\textbf{49.3}&\multicolumn{1}{c|}{45.1}&\underline{\textbf{58.6}}&45.1\\
        \multicolumn{1}{c|}{}&\multicolumn{1}{c|}{PGD-$l_\infty$} &
            0.0&0.0&\multicolumn{1}{c|}{\underline{\textbf{31.6}}}&22.1&\textbf{25.4}\\
        \multicolumn{1}{c|}{}&\multicolumn{1}{c|}{PGD-$l_2$} &
            0.1&1.5&\multicolumn{1}{c|}{\textbf{46.8}}&40.4&\underline{\textbf{48.7}}\\
        \cline{2-7}
        \multicolumn{1}{c|}{}&\multicolumn{1}{c|}{Corrupted-1} &   
        89.5&\underline{\textbf{91.7}}&\multicolumn{1}{c|}{87.3}&\textbf{89.8}&88.3\\
        \multicolumn{1}{c|}{}&\multicolumn{1}{c|}{Corrupted-2} &   
        87.1&\underline{\textbf{90.3}}&\multicolumn{1}{c|}{86.1}&\textbf{88.3}&87.1\\
        \multicolumn{1}{c|}{}&\multicolumn{1}{c|}{Corrupted-3} &   
        85.0&\underline{\textbf{89.0}}&\multicolumn{1}{c|}{85.1}&\textbf{87.0}&85.9\\
        \multicolumn{1}{c|}{}&\multicolumn{1}{c|}{Corrupted-4} &   
        82.2&\underline{\textbf{87.1}}&\multicolumn{1}{c|}{83.6}&\textbf{85.2}&84.3\\
        \multicolumn{1}{c|}{}&\multicolumn{1}{c|}{Corrupted-5} &   
        78.3&\underline{\textbf{84.5}}&\multicolumn{1}{c|}{81.1}&\textbf{82.3}&81.7\\
        \hline
    \end{tabular}
\end{table}
\setlength{\tabcolsep}{7pt}
\begin{table}[t]
    \centering
    \caption{The classification accuracy  (\%) of WideResNet40-2 classifiers trained on CIFAR-100 with different combination of images. The FGSM attack with $\epsilon=8/255$ was used in training. The top-2 results are indicated in bold while the best results are underlined.}
    \label{table:cifar100_wrn_fgsm8}
    \begin{tabular}{cccccc}
        \hline
        \multicolumn{1}{c|}{}& \multicolumn{5}{c}{Combination of Training Data} \\
        \multicolumn{1}{c|}{}&Clean&APR&\multicolumn{1}{c|}{C\&Adv}&C\&AA&C\&AP\\
        \hline
        \multicolumn{1}{c|}{Clean} &   
            \underline{\textbf{72.5}}&\textbf{72.0}&\multicolumn{1}{c|}{63.7}&66.3&61.3\\
        \hline
        \multicolumn{1}{c|}{FGSM ($\epsilon_0=4$)} &
            32.8&35.5&\multicolumn{1}{c|}{\underline{\textbf{67.8}}}&\textbf{43.9}&36.5\\
        \multicolumn{1}{c|}{FGSM ($\epsilon_0=8$)} &
            27.5&31.7&\multicolumn{1}{c|}{\underline{\textbf{70.1}}}&\textbf{38.8}&31.1\\
        \multicolumn{1}{c|}{FGSM ($\epsilon_0=16$)} &
            20.5&27.2&\multicolumn{1}{c|}{\underline{\textbf{55.1}}}&\textbf{33.1}&26.4\\
        \multicolumn{1}{c|}{FGSM ($\epsilon_0=32$)} &
            12.7&20.4&\multicolumn{1}{c|}{\underline{\textbf{41.9}}}&\textbf{27.3}&21.1\\
        \multicolumn{1}{c|}{PGD-$l_\infty$} &
            0.0&0.0&\multicolumn{1}{c|}{1.3}&\textbf{8.9}&\underline{\textbf{13.3}}\\
        \multicolumn{1}{c|}{PGD-$l_2$} &
            0.1&1.0&\multicolumn{1}{c|}{0.5}&\textbf{19.5}&\underline{\textbf{24.3}}\\
        \hline
        \multicolumn{1}{c|}{Corrupted-1} &   
        \textbf{70.3}&\underline{\textbf{73.5}}&\multicolumn{1}{c|}{67.2}&68.9&66.2\\
        \multicolumn{1}{c|}{Corrupted-2} &   
        66.8&\underline{\textbf{71.7}}&\multicolumn{1}{c|}{65.4}&\textbf{67.0}&64.6\\
        \multicolumn{1}{c|}{Corrupted-3} &   
        63.9&\underline{\textbf{69.8}}&\multicolumn{1}{c|}{63.5}&\textbf{65.4}&62.9\\
        \multicolumn{1}{c|}{Corrupted-4} &   
        60.3&\underline{\textbf{67.2}}&\multicolumn{1}{c|}{61.1}&\textbf{63.3}&61.1\\
        \multicolumn{1}{c|}{Corrupted-5} &   
        55.2&\underline{\textbf{63.4}}&\multicolumn{1}{c|}{57.5}&\textbf{59.7}&57.9\\
        \hline
    \end{tabular}
\end{table}

Table~\ref{table:cifar10_fgsm8} shows the results of three different networks trained on the CIFAR-10 dataset with different combinations of training images. The FGSM attack with $\epsilon=8/255$ was used in training. For ResNet-18, the model trained with the clean images (Clean) achieved the best performance on clean accuracy. The model trained with the APR images (APR) achieved the best performance in terms of robustness against common corruptions. However, these models were vulnerable to adversarial perturbations, particularly to those by the PGD. 
The model trained with clean and FGSM adversarial images (C\&Adv) was the most robust against FGSM perturbations. The model trained with clean and adversarial amplitude images (C\&AA) achieved an overall improvement in the robustness test compared to the model trained with clean images. Compared to the APR images, the adversarial images greatly improved the robustness of models against adversarial perturbations. Note that the adversarial phase images (C\&AP) also improved the model robustness against adversarial perturbations but failed to improve the model robustness against common corruptions, compared to the model trained with only clean images (Clean). The model tended to focus on learning the information included in the amplitude spectrum and hence became more sensitive to common corruptions. 

The models with bigger network architectures, e.g., WideResNet-40-2 and DenseNet, were capable of learning from FGSM adversarial images to be robust against PGD perturbations when adversarial images were used in training (C\&Adv). However, simultaneously, the model became more vulnerable to common corruptions. In contrast, the model trained with adversarial amplitude images (C\&AA) became more robust to both common corruptions and adversarial perturbations compared to the model trained with only clean images (Clean). 

Comparing the results across different networks, we note that the network with a smaller architecture, i.e., the ResNet-18, struggled to learn directly from FGSM adversarial images. This led to catastrophic overfitting of FGSM perturbations and a failure to be robust against PGD perturbations. With the proposed method, both the adversarial amplitude images and adversarial phase images helped the model to learn these adversarial features without overfitting the FGSM perturbations. In contrast, the models with larger architectures, that is, the WideResNet and the DenseNet did not overfit the FGSM perturbations. Both of these networks could learn from the FGSM images and became robust against PGD perturbations when trained with adversarial images. In networks with larger architectures, adversarial amplitude images improved both the model robustness against common corruptions and adversarial perturbations. 

We further evaluated the methods on the CIFAR-100 dataset using the same experimental setup, as shown in~Table~\ref{table:cifar100_wrn_fgsm8}. Similar results were observed, in which the adversarial amplitude images improved robustness against both common corruptions and the adversarial images. Refer to the supplementary material for more results on the CIFAR-100 and the ImageNet~\cite{deng2009imagenet} datasets. 

To further evaluate the proposed method, we used PGD-$l_\infty$ attack with $\epsilon=8/255, \alpha=0.1, i_\text{iters}=10$ in training. Table~\ref{table:cifar10_pgd} shows the classification accuracy of models trained under the same experimental setup for WideResNet-40-2 classifiers. The results indicate that the model trained with adversarial amplitude images (C\&AA) outperformed the model trained with adversarial images (C\&Adv) in almost all of the aspects including clean accuracy, adversarial robustness, and robustness against common corruptions. The model (C\&AA) also improved both the adversarial robustness and common corruptions compared to the model trained with clean images (Clean). In short, the training with adversarial amplitude images led to CNN classifiers that are robust to both common corruptions and adversarial perturbations. The results for other models and the CIFAR-100 dataset can be found in the supplementary material. 
\setlength{\tabcolsep}{7pt}
\begin{table}[t]
    \centering
    \caption{The classification accuracy  (\%) of WideResNet-40-2 classifiers trained on CIFAR-10 with different combination of images. The PGD-$l_\infty$ attack with $\epsilon=8/255, \alpha=2/255, i_\text{iters}=10$ was used in training. The top-2 results are indicated in bold, and the best results are underlined.}
    \label{table:cifar10_pgd}
    \begin{tabular}{ccccccc}
        \hline
        & \multicolumn{1}{c|}{}& \multicolumn{5}{c}{Combination of Training Data} \\
        &\multicolumn{1}{c|}{}&Clean&APR&\multicolumn{1}{c|}{C\&Adv}&C\&AA&C\&AP\\
        \hline
        \multicolumn{1}{c|}{\multirow{11}{*}{WideResNet}}&\multicolumn{1}{c|}{Clean} &   
            \textbf{94.1}&\underline{\textbf{94.3}}&\multicolumn{1}{c|}{83.8}&88.5&86.2\\
        \cline{2-7}
        \multicolumn{1}{c|}{}&\multicolumn{1}{c|}{FGSM ($\epsilon_0=4$)} &
            \underline{\textbf{72.5}}&69.7&\multicolumn{1}{c|}{65.7}&\textbf{70.9}&68.2\\
        \multicolumn{1}{c|}{}&\multicolumn{1}{c|}{FGSM ($\epsilon_0=8$)} &
            \underline{\textbf{66.8}}&64.5&\multicolumn{1}{c|}{55.2}&\textbf{66.2}&62.8\\
        \multicolumn{1}{c|}{}&\multicolumn{1}{c|}{FGSM ($\epsilon_0=16$)} &
            58.8&\textbf{60.0}&\multicolumn{1}{c|}{45.2}&\underline{\textbf{61.4}}&57.5\\
        \multicolumn{1}{c|}{}&\multicolumn{1}{c|}{FGSM ($\epsilon_0=32$)} &
            43.5&41.0&\multicolumn{1}{c|}{38.3}&\underline{\textbf{56.7}}&\textbf{51.9}\\
        \multicolumn{1}{c|}{}&\multicolumn{1}{c|}{PGD-$l_\infty$} &
        0.2&0.3&\multicolumn{1}{c|}{\underline{\textbf{46.2}}}&\textbf{42.6}&39.6\\
        \multicolumn{1}{c|}{}&\multicolumn{1}{c|}{PGD-$l_2$} &
            2.4&7.0&\multicolumn{1}{c|}{\textbf{52.1}}&\underline{\textbf{52.7}}&51.6\\
        \cline{2-7}
        \multicolumn{1}{c|}{}&\multicolumn{1}{c|}{Corrupted-1} &   
            \textbf{90.4}&\underline{\textbf{92.2}}&\multicolumn{1}{c|}{84.5}&88.0&86.4\\
        \multicolumn{1}{c|}{}&\multicolumn{1}{c|}{Corrupted-2} &   
            \textbf{88.1}&\underline{\textbf{90.9}}&\multicolumn{1}{c|}{83.3}&86.9&85.3\\
        \multicolumn{1}{c|}{}&\multicolumn{1}{c|}{Corrupted-3} &   
            \textbf{85.9}&\underline{\textbf{89.7}}&\multicolumn{1}{c|}{82.2}&85.7&84.1\\
        \multicolumn{1}{c|}{}&\multicolumn{1}{c|}{Corrupted-4} &   
            83.2&\underline{\textbf{87.7}}&\multicolumn{1}{c|}{80.7}&\textbf{84.1}&82.5\\
        \multicolumn{1}{c|}{}&\multicolumn{1}{c|}{Corrupted-5} &   
            79.3&\underline{\textbf{85.1}}&\multicolumn{1}{c|}{77.7}&\textbf{81.3}&79.7\\
        \hline 
    \end{tabular}
\end{table}

\subsection{Ablation Study}\label{sec:experiment-ablation}
With the proposed data augmentation method, four types of images can be used in training, including clean images (Clean), FGSM adversarial images (Adv), adversarial amplitude images (AA), and adversarial phase images (AP). Note that the training with adversarial images corresponds to Madry's adversarial training~\cite{madry2018towards}.
We investigated the effect of each type of image on the generalization of WideResNet-40-2 classifiers. Table~\ref{table:ablation} shows the results. When only adversarial images are used in training, the model overfitted the FGSM adversarial perturbations and hence failed to produce a similar performance on the clean and corrupted images, as well as the PGD adversarial images. With the adversarial amplitude images, the model can learn to be robust against both FGSM and PGD perturbations. Because the amplitude spectrum is replaced by that of adversarial images, the model is encouraged to learn more semantic information included in the phase spectrum of clean images. Simultaneously, the adversarially perturbed amplitude spectrum encourages the model to learn adversarial features. Hence, a model trained with adversarial amplitude images becomes more robust against both common corruptions and adversarial perturbations. 
\setlength{\tabcolsep}{7pt}
\begin{table}[t]
    \centering
    \caption{The classification accuracy  (\%) of WideResNet40-2 classifiers trained on CIFAR-10 with different type of images. The FGSM attack with $\epsilon=8/255$ was used in training. The top-2 results are indicated in bold, and the best results are underlined.}
    \label{table:ablation}
    \begin{tabular}{cccccc}
        \hline
        \multicolumn{1}{c|}{}&\multicolumn{1}{c|}{}& \multicolumn{4}{c}{Training Images} \\
        \multicolumn{1}{c|}{}&\multicolumn{1}{c|}{}&Clean&Adv&\multicolumn{1}{|c}{AA}&AP\\
        \hline
        \multicolumn{1}{c|}{\multirow{11}{*}{CIFAR-10}}&\multicolumn{1}{c|}{Clean} &   
            \underline{\textbf{94.1}}&\multicolumn{1}{c|}{86.8}&\textbf{89.0}&86.9\\
        \cline{2-6}
        \multicolumn{1}{c|}{}&\multicolumn{1}{c|}{FGSM ($\epsilon_0=4$)} &
            72.5&\multicolumn{1}{c|}{\underline{\textbf{88.7}}}&\textbf{74.7}&71.2\\
        \multicolumn{1}{c|}{}&\multicolumn{1}{c|}{FGSM ($\epsilon_0=8$)} &
            66.8&\multicolumn{1}{c|}{\underline{\textbf{88.6}}}&\textbf{70.3}&66.8\\
        \multicolumn{1}{c|}{}&\multicolumn{1}{c|}{FGSM ($\epsilon_0=16$)} &
            58.8&\multicolumn{1}{c|}{\underline{\textbf{87.6}}}&\textbf{65.3}&61.6\\
        \multicolumn{1}{c|}{}&\multicolumn{1}{c|}{FGSM ($\epsilon_0=32$)} &
            43.5&\multicolumn{1}{c|}{\underline{\textbf{85.3}}}&\textbf{59.8}&54.9\\
        \multicolumn{1}{c|}{}&\multicolumn{1}{c|}{PGD-$l_\infty$} &
            0.2&\multicolumn{1}{c|}{19.1}&\textbf{40.4}&\underline{\textbf{40.8}}\\
        \multicolumn{1}{c|}{}&\multicolumn{1}{c|}{PGD-$l_2$} &
            2.4&\multicolumn{1}{c|}{12.3}&\textbf{53.5}&\underline{\textbf{53.6}}\\
        \cline{2-6}
        \multicolumn{1}{c|}{}&\multicolumn{1}{c|}{Corrupted-1} &   
        \underline{\textbf{90.4}}&\multicolumn{1}{c|}{86.5}&\textbf{88.4}&86.7\\
        \multicolumn{1}{c|}{}&\multicolumn{1}{c|}{Corrupted-2} &   
        \underline{\textbf{88.1}}&\multicolumn{1}{c|}{85.1}&\textbf{87.2}&85.5\\
        \multicolumn{1}{c|}{}&\multicolumn{1}{c|}{Corrupted-3} &   
        \textbf{85.9}&\multicolumn{1}{c|}{83.5}&\underline{\textbf{86.1}}&84.4\\
        \multicolumn{1}{c|}{}&\multicolumn{1}{c|}{Corrupted-4} &   
        \textbf{83.2}&\multicolumn{1}{c|}{81.6}&\underline{\textbf{84.4}}&82.8\\
        \multicolumn{1}{c|}{}&\multicolumn{1}{c|}{Corrupted-5} &   
        79.3&\multicolumn{1}{c|}{78.0}&\underline{\textbf{81.5}}&\textbf{79.8}\\
        \hline
    \end{tabular}
\end{table}

\subsection{Extreme Cases}\label{sec:experiment-extreme}
\begin{figure}[t]
    \centering
    \includegraphics[width=0.95\textwidth]{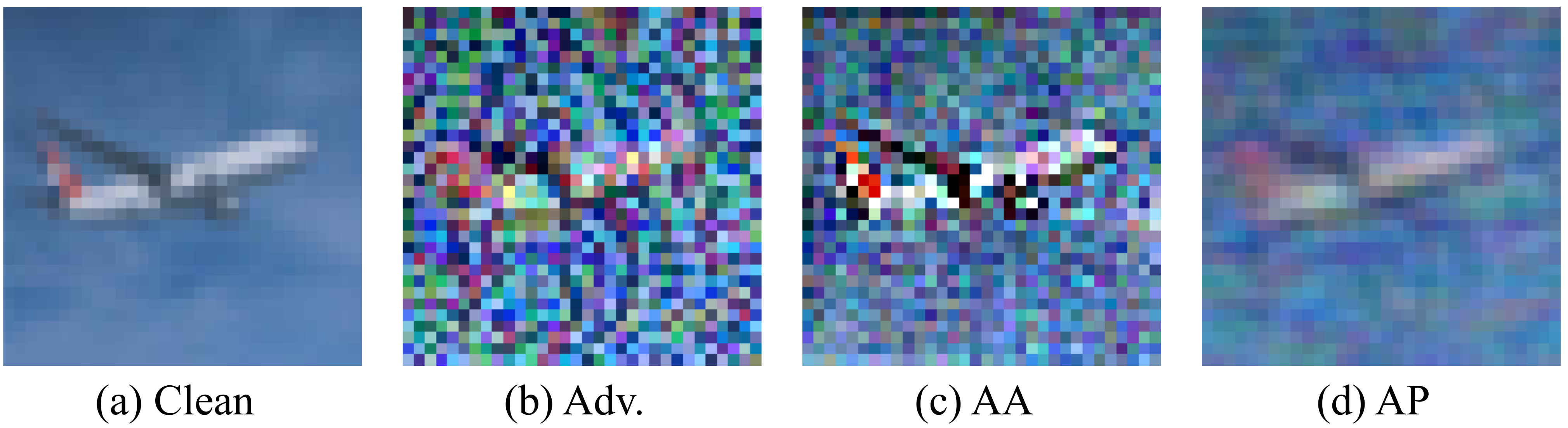}
    \caption{Examples of images generated by the PGD-$l_\infty$ attack with $\epsilon=0.5, \alpha=0.1, i_\text{iters}=10$. (a) clean image, (b) adversarial image, (c) adversarial amplitude image, and (d) adversarial phase image. }
    \label{fig:extreme}
\end{figure}
\setlength{\tabcolsep}{7pt}
\begin{table}[t]
    \centering
    \caption{The classification accuracy  (\%) of WideResNet-40-2 classifiers trained on CIFAR-10. An extremely strong PGD-$l_\infty$ attack with $\epsilon=0.5, \alpha=0.1, i_\text{iters}=10$ was used in training. The best results are indicated in bold.}
    \label{table:extreme}
    \begin{tabular}{cccccc}
        \hline
        \multicolumn{1}{c|}{}& \multicolumn{5}{c}{Combination of Training Data}\\
        \multicolumn{1}{c|}{}&\multicolumn{1}{c|}{Clean\&Adv}&Clean\&AA&Clean\&AP\\
        \hline
        \multicolumn{1}{c|}{Clean} &   
            \multicolumn{1}{c|}{90.0}&\textbf{93.7}&89.9\\
        \hline
        \multicolumn{1}{c|}{FGSM ($\epsilon_0=4$)} &
            \multicolumn{1}{c|}{61.2}&\textbf{74.0}&62.2\\
        \multicolumn{1}{c|}{FGSM ($\epsilon_0=8$)} &
            \multicolumn{1}{c|}{33.6}&\textbf{71.1}&61.3\\
        \multicolumn{1}{c|}{FGSM ($\epsilon_0=16$)} &
            \multicolumn{1}{c|}{21.7}&\textbf{69.8}&55.6\\
        \multicolumn{1}{c|}{FGSM ($\epsilon_0=32$)} &
            \multicolumn{1}{c|}{21.3}&\textbf{64.0}&53.9\\
        \multicolumn{1}{c|}{PGD-$l_\infty$} &
            \multicolumn{1}{c|}{0.0}&\textbf{2.1}&0.1\\
        \multicolumn{1}{c|}{PGD-$l_2$} &
            \multicolumn{1}{c|}{0.1}&\textbf{15.5}&0.9\\
        \hline
        \multicolumn{1}{c|}{Corrupted-1} &   
        \multicolumn{1}{c|}{79.7}&\textbf{90.6}&83.6\\
        \multicolumn{1}{c|}{Corrupted-2} &   
        \multicolumn{1}{c|}{72.7}&\textbf{88.7}&80.4\\
        \multicolumn{1}{c|}{Corrupted-3} &   
        \multicolumn{1}{c|}{71.5}&\textbf{87.0}&78.5\\
        \multicolumn{1}{c|}{Corrupted-4} &   
        \multicolumn{1}{c|}{67.7}&\textbf{84.5}&76.2\\
        \multicolumn{1}{c|}{Corrupted-5} &   
        \multicolumn{1}{c|}{63.5}&\textbf{80.2}&71.9\\
        \hline
    \end{tabular}
\end{table}
We also trained WideResNet40-2 classifiers on CIFAR-10 with an extremely strong adversarial attack. 
We trained the models using a combination of clean images and either of the other options. The PGD-$l_\infty$ attack with $\epsilon=0.5, \alpha=0.1, i_\text{iters}=10$, which allows the adversary to change at most 50\% of the target image, was used in training.
These adversarial images can barely be recognized by humans. However, with the adversarial amplitude swap, the adversarial amplitude and adversarial phase images are more recognizable than the adversarial images (Fig.~\ref{fig:extreme}). Table~\ref{table:extreme} shows the results. When adversarial images were used, the model struggled to learn adversarial features, and hence it failed to be robust against adversarial perturbations. In contrast, adversarial amplitude images, which contain only the adversarially designed amplitude spectrum, enable the model to learn to be robust against adversarial perturbations. 
Both the models trained with adversarial amplitude and adversarial phase images were more robust against adversarial perturbations. 

\subsection{Catastrophic Overfitting and Robust Overfitting}~\label{sec:experiment-overfitting}
\begin{figure}[t]
    \centering
    \includegraphics[width=0.95\textwidth]{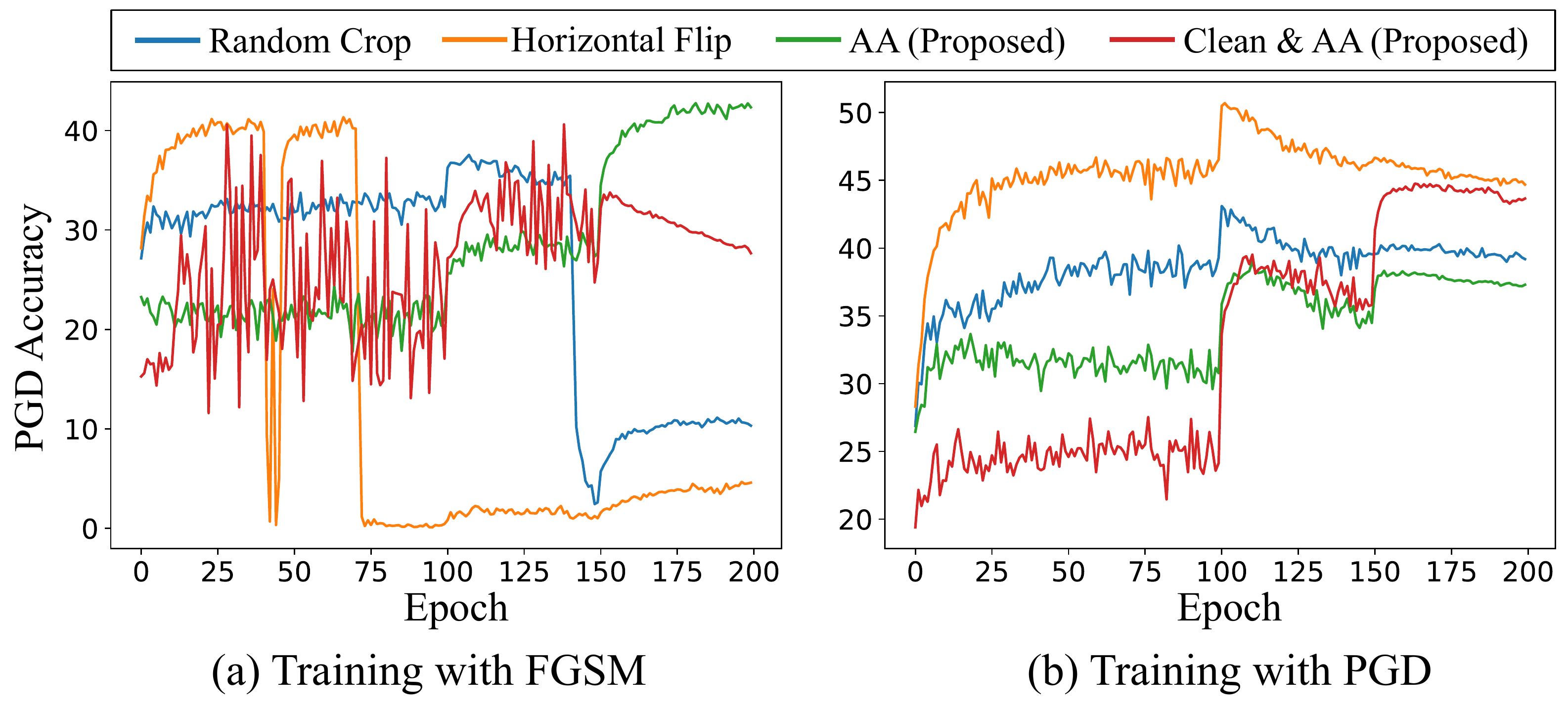}
    \caption{WideResNet-40-2 classifiers trained with conventional and proposed data augmentation methods. (a) Training with FGSM perturbations. (b) Training with PGD perturbations. }
    \label{fig:overfitting}
\end{figure}

\citet{tack2021consistency} revealed that data augmentation methods are effective in preventing robust overfitting during adversarial training. We compared the proposed method with two conventional data augmentation methods, including random crop and random horizontal flip. 
Figure~\ref{fig:overfitting} shows the model robustness against PGD perturbations when WideResNet-40-2 classifiers are trained using different data augmentation methods on the FGSM and the PGD perturbed images. 
Both the models trained with either of the conventional data augmentation methods suffered from both catastrophic and robust overfitting. 
When FGSM perturbed images were used in training, the model trained with adversarial amplitude swap (AA) did not suffer from catastrophic overfitting, leading to improvements in accuracy at the 100th and 150th epochs, and eventually outperformed both the conventional data augmentation methods. This indicates that adversarial amplitude images contain moderate perturbations, which allows the classifiers to learn from them more easily. 
When PGD perturbed images are used in training, the models trained with random cropped images, horizontal flipped images, and also the adversarial amplitude images suffered from robust overfitting, and hence failed to deliver improvements in PGD accuracy after the learning rates are tuned at the 100th and 150th epochs. However, when the clean and adversarial amplitude images (Clean \& AA) were used in training, the model did not suffer from robust overfitting. Although it started at a relatively low accuracy during the first 100 epochs, the model could pick up the adversarial features from the amplitude spectrum after the tuning of learning rates at the 100th and 150th epochs. 

We noticed a few interesting phenomena from the results. When FGSM perturbed images were used in training, the model performed better when it was trained with adversarial amplitude images (AA). On the other side, when PGD perturbed images were used in training, the model performed better when it was trained with both the clean and adversarial amplitude images (Clean \& AA). These results encouraged rethinking of methods to prepare training images from the frequency perspective in the adversarial training setup. Despite the differences in the performance of models trained with or without clean images, the proposed method, which includes both the clean and adversarial amplitude images improved the model robustness against both common corruptions and adversarial perturbations, as shown in Section~\ref{sec:experiment-classification}. 

\subsection{Adversarial Features in Adversarial Amplitude Images}
We demonstrated the efficacy of the proposed method in training a robust classifier against both common corruptions and adversarial perturbations. Notably, despite the ability to improve adversarial robustness, adversarial amplitude images did not serve as a strong adversarial attack compared to original adversarial images, as shown in Table~\ref{table:errorrate}. We tested several classifiers trained with clean images and showed the classification errors of the classifiers on clean, adversarial (Adv), adversarial amplitude (AA), and adversarial phase (AP) images. We noticed that the original adversarial images served as the strongest attack. Because adversarial amplitude and adversarial phase images contain only either one of the frequency spectra of adversarial images, both of these images were expected to have a lower fooling rate on the classifiers. The results indicate that the adversarial phase images, which contain the phase spectrum of adversarial images, are better in fooling the classifiers, compared to adversarial amplitude images. Nevertheless, the CNN classifiers trained with adversarial amplitude images showed comparable or, in some cases, even better adversarial robustness compared to the standard adversarial training. 

\setlength{\tabcolsep}{10pt}
\begin{table}[t]
    \centering
    \caption{Classification error of classifiers trained with clean images on several types of adversarially perturbed images.}
    \label{table:errorrate}
    \begin{tabular}{cccccc}
        \hline
        \multicolumn{1}{c|}{}&\multicolumn{1}{c|}{\multirow{2}{*}{Networks}}& \multicolumn{4}{c}{Error Rates (\%)} \\
        \multicolumn{1}{c|}{}&\multicolumn{1}{c|}{}&Clean&Adv&AA&AP\\
        \hline
        \multicolumn{1}{c|}{\multirow{3}{*}{FGSM}}&\multicolumn{1}{c|}{ResNet-18} &   
            5.2&34.4&22.6&27.0\\
        \multicolumn{1}{c|}{}&\multicolumn{1}{c|}{WideResNet} &
            6.0&33.2&21.5&26.3\\
        \multicolumn{1}{c|}{}&\multicolumn{1}{c|}{DenseNet} &
            6.6&42.7&29.6&34.8\\
        \hline
        \multicolumn{1}{c|}{\multirow{3}{*}{PGD-$l_\infty$}}&\multicolumn{1}{c|}{ResNet-18} &   
            5.2&99.9&37.8&76.4\\
        \multicolumn{1}{c|}{}&\multicolumn{1}{c|}{WideResNet} &
            6.0&99.8&36.0&76.5\\
        \multicolumn{1}{c|}{}&\multicolumn{1}{c|}{DenseNet} &
            6.6&100.0&50.4&87.38\\
    \end{tabular}
\end{table}

\section{Conclusion}
In this study, we have investigated the nature of the amplitude and phase spectra of adversarial images. Inspired by previous works on altering the frequency spectra of training images, we have proposed a frequency-based data augmentation method, in which the amplitude spectrum is swapped between clean and adversarial images. 
We have demonstrated that training with adversarial amplitude images led to CNN classifiers that are robust against both common corruptions and adversarial perturbations. The experimental results have shown that the amplitude spectrum of adversarial images accommodates moderate and general perturbations that can help CNN classifiers to equip more general robustness, while the phase spectrum of adversarial images is moderate but still adversarial, which may improve the adversarial robustness of classifiers but also retains the risk of suffering from catastrophic and robust overfitting. Despite having a weaker fooling ability, adversarial amplitude images served as better training images that helped CNN classifiers achieve more general robustness. We further demonstrated that with the proposed data augmentation method, CNNs can even learn from some extreme adversarial examples that humans can barely recognize.  
We believe that these findings will be crucial for the future development of truly robust classifiers. 

\renewcommand\bibname{\leftline{References}}
\bibliographystyle{splncs04nat}

\end{document}


\maketitle

This supplementary material presents more results of the experiments. Cross-referencing numbers here are prefixed with S (e.g., Table.~S1 or Fig.~S1). Numbers without the prefix (e.g., Table.~1 or Fig.~1) refer to numbers in the main text.

\section{CIFAR-10 \& CIFAR-100 Image Classification}\label{sec:cifar}
In addition to the results in Section~\cifar{}, we present a complete result on the classification accuracy of models trained on CIFAR-10 and CIFAR-100 datasets~\cite{krizhevsky2009multilayersfeatures}. 
Table~\ref{table:cifar100-fgsm} shows the classification accuracy of models trained on the CIFAR-100 dataset with FGSM used in training. The general trend of the results is the same as the models trained on the CIFAR-10 dataset~(Table~\cifartenfgsm{}), where the models trained with clean and adversarial amplitude images are robust to both common corruptions and adversarial perturbations while those trained with APR or adversarial images only improved one of those aspects. 
Models of WideResNet did not overfit FGSM perturbation when they were trained on CIFAR-10 but they overfitted when they were trained on the CIFAR-100 dataset. On the contrary, models of ResNet-18 overfitted when they were trained on the CIFAR-10 dataset but did not overfit when they were trained on the CIFAR-100 dataset. This indicates that CNN classifiers struggled to generalize well on adversarial images. In contrast, adversarial amplitude images served as better images for classifiers to learn to be robust against adversarial perturbations as they contain moderate and general perturbations compared to adversarial images. Regarding the robustness against common corruptions, both APR images and adversarial amplitude images improved the model robustness against common corruptions compared to those trained with clean images. 

Tables~\ref{table:cifar10_pgd} and~\ref{table:cifar100_pgd} show the complete results of models with three different network architectures trained on CIFAR-10 and CIFAR-100 datasets when PGD was used in training. The result of WideResNet models trained on CIFAR-10 shown in Table~\cifartenpgd{} is also included in Table~\ref{table:cifar10_pgd}. When PGD adversarial images were used in training, all the models could learn to be robust against adversarial images without overfitting FGSM perturbations. However, we also noticed a degradation in robustness against FGSM perturbations despite the improvement in robustness against PGD perturbations. In contrast, the models trained with adversarial amplitude images produced the highest general robustness against adversarial perturbations. This indicates that adversarial amplitude images helped the classifiers learn adversarial features more easily. Regarding the robustness against common corruption, the results are similar to those that were trained with the FGSM, where both the APR and adversarial amplitude images improved the robustness against common corruptions compared to those that were trained with clean images. 

We conducted the same experiment by training models with ResNet-18 architecture on Tiny ImageNet~\cite{le2015tiny}. Tiny ImageNet is a subset of ImageNet~\cite{deng2009imagenet}, which contains 100,000 images of 200 classes downsized to 64x64 resolution. Each class consists of 500 training images and 50 test images. With this dataset, we fine-tuned ResNet-18 models pretrained on the ImageNet dataset, which are provided in PyTorch library~\cite{paszke2019pytorch}. FGSM was used to generate adversarial perturbations in training. 
The evaluation was done on the validation dataset. The results are shown in Table~\ref{table:imagenet}. Similar to the results on CIFAR-10 and CIFAR-100, adversarial amplitude images improved the robustness against both common corruptions and adversarial perturbations compared to models trained with clean images. 
In this training setup, APR images produced higher robustness against corruptions with the severity of levels 1--2, while adversarial amplitude images produced higher robustness against corruptions with the severity of levels 3--5.
In addition, adversarial amplitude images also improved model robustness against adversarial perturbations; however, this was not the case for APR images. This indicates that adversarial amplitude images lead to CNN classifiers with more general robustness. 

\setlength{\tabcolsep}{7pt}
\begin{table}[p]
    \centering
    \caption{The classification accuracy (\%) of ResNet-18, WideResNet-40-2, and DenseNet classifiers trained on CIFAR-100 with different combination of images. The FGSM attack with $\epsilon=8/255$ was used in training. The top-2 results are indicated in bold while the best results are underlined.}
    \label{table:cifar100-fgsm}
    \begin{tabular}{ccccccc}
        \\
        \hline
        & \multicolumn{1}{c|}{}& \multicolumn{5}{c}{Combination of Training Data} \\
        &\multicolumn{1}{c|}{}&Clean&APR&\multicolumn{1}{c|}{C\&Adv}&C\&AA&C\&AP\\
        \hline
        \multicolumn{1}{c|}{\multirow{11}{*}{ResNet-18}}&\multicolumn{1}{c|}{Clean} &   
            \underline{\textbf{75.9}}&\textbf{74.5}&\multicolumn{1}{c|}{60.5}&70.0&64.1\\
        \cline{2-7}
        \multicolumn{1}{c|}{}&\multicolumn{1}{c|}{FGSM ($\epsilon_0=4$)} &
            36.6&36.9&\multicolumn{1}{c|}{39.6}&\underline{\textbf{48.2}}&\textbf{41.8}\\
        \multicolumn{1}{c|}{}&\multicolumn{1}{c|}{FGSM ($\epsilon_0=8$)} &
            30.3&32.7&\multicolumn{1}{c|}{33.8}&\underline{\textbf{43.8}}&\textbf{36.7}\\
        \multicolumn{1}{c|}{}&\multicolumn{1}{c|}{FGSM ($\epsilon_0=16$)} &
            24.0&29.2&\multicolumn{1}{c|}{29.2}&\underline{\textbf{38.2}}&\textbf{31.2}\\
        \multicolumn{1}{c|}{}&\multicolumn{1}{c|}{FGSM ($\epsilon_0=32$)} &
            16.4&23.1&\multicolumn{1}{c|}{24.4}&\underline{\textbf{31.7}}&\textbf{24.8}\\
        \multicolumn{1}{c|}{}&\multicolumn{1}{c|}{PGD-$l_\infty$} &
        0.0&0.0&\multicolumn{1}{c|}{\underline{\textbf{20.7}}}&12.6&\textbf{16.6}\\
        \multicolumn{1}{c|}{}&\multicolumn{1}{c|}{PGD-$l_2$} &
        0.2&0.4&\multicolumn{1}{c|}{29.1}&\underline{\textbf{29.3}}&\textbf{31.0}\\
        \cline{2-7}
        \multicolumn{1}{c|}{}&\multicolumn{1}{c|}{Corrupted-1} &   
            \textbf{74.3}&\underline{\textbf{75.9}}&\multicolumn{1}{c|}{65.7}&72.2&68.6\\
        \multicolumn{1}{c|}{}&\multicolumn{1}{c|}{Corrupted-2} &   
            \textbf{71.3}&\underline{\textbf{74.2}}&\multicolumn{1}{c|}{64.3}&70.6&67.2\\
        \multicolumn{1}{c|}{}&\multicolumn{1}{c|}{Corrupted-3} &   
            68.7&\underline{\textbf{72.6}}&\multicolumn{1}{c|}{63.0}&\textbf{69.1}&65.7\\
        \multicolumn{1}{c|}{}&\multicolumn{1}{c|}{Corrupted-4} &   
            65.6&\underline{\textbf{70.2}}&\multicolumn{1}{c|}{61.3}&\textbf{67.0}&64.1\\
        \multicolumn{1}{c|}{}&\multicolumn{1}{c|}{Corrupted-5} &   
            61.0&\underline{\textbf{66.9}}&\multicolumn{1}{c|}{58.5}&\textbf{63.9}&61.3\\
        \hline 
        \hline
        \multicolumn{1}{c|}{\multirow{11}{*}{WideResNet}}&\multicolumn{1}{c|}{Clean} &   
            \underline{\textbf{72.5}}&\textbf{72.0}&\multicolumn{1}{c|}{63.7}&66.3&61.3\\
        \hline
        \multicolumn{1}{c|}{}&\multicolumn{1}{c|}{FGSM ($\epsilon_0=4$)} &
            32.8&35.5&\multicolumn{1}{c|}{\underline{\textbf{67.8}}}&\textbf{43.9}&36.5\\
        \multicolumn{1}{c|}{}&\multicolumn{1}{c|}{FGSM ($\epsilon_0=8$)} &
            27.5&31.7&\multicolumn{1}{c|}{\underline{\textbf{70.1}}}&\textbf{38.8}&31.1\\
        \multicolumn{1}{c|}{}&\multicolumn{1}{c|}{FGSM ($\epsilon_0=16$)} &
            20.5&27.2&\multicolumn{1}{c|}{\underline{\textbf{55.1}}}&\textbf{33.1}&26.4\\
        \multicolumn{1}{c|}{}&\multicolumn{1}{c|}{FGSM ($\epsilon_0=32$)} &
            12.7&20.4&\multicolumn{1}{c|}{\underline{\textbf{41.9}}}&\textbf{27.3}&21.1\\
        \multicolumn{1}{c|}{}&\multicolumn{1}{c|}{PGD-$l_\infty$} &
            0.0&0.0&\multicolumn{1}{c|}{1.3}&\textbf{8.9}&\underline{\textbf{13.3}}\\
        \multicolumn{1}{c|}{}&\multicolumn{1}{c|}{PGD-$l_2$} &
            0.1&1.0&\multicolumn{1}{c|}{0.5}&\textbf{19.5}&\underline{\textbf{24.3}}\\
        \cline{2-7}
        \multicolumn{1}{c|}{}&\multicolumn{1}{c|}{Corrupted-1} &   
        \textbf{70.3}&\underline{\textbf{73.5}}&\multicolumn{1}{c|}{67.2}&68.9&66.2\\
        \multicolumn{1}{c|}{}&\multicolumn{1}{c|}{Corrupted-2} &   
        66.8&\underline{\textbf{71.7}}&\multicolumn{1}{c|}{65.4}&\textbf{67.0}&64.6\\
        \multicolumn{1}{c|}{}&\multicolumn{1}{c|}{Corrupted-3} &   
        63.9&\underline{\textbf{69.8}}&\multicolumn{1}{c|}{63.5}&\textbf{65.4}&62.9\\
        \multicolumn{1}{c|}{}&\multicolumn{1}{c|}{Corrupted-4} &   
        60.3&\underline{\textbf{67.2}}&\multicolumn{1}{c|}{61.1}&\textbf{63.3}&61.1\\
        \multicolumn{1}{c|}{}&\multicolumn{1}{c|}{Corrupted-5} &   
        55.2&\underline{\textbf{63.4}}&\multicolumn{1}{c|}{57.5}&\textbf{59.7}&57.9\\
        \hline
        \hline
        \multicolumn{1}{c|}{\multirow{11}{*}{DenseNet}}&\multicolumn{1}{c|}{Clean} &   
            \textbf{71.3}&\underline{\textbf{71.5}}&\multicolumn{1}{c|}{60.6}&68.8&62.5\\
        \cline{2-7}
        \multicolumn{1}{c|}{}&\multicolumn{1}{c|}{FGSM ($\epsilon_0=4$)} &
            28.3&27.7&\multicolumn{1}{c|}{\textbf{34.3}}&\underline{\textbf{41.8}}&30.4\\
        \multicolumn{1}{c|}{}&\multicolumn{1}{c|}{FGSM ($\epsilon_0=8$)} &
            24.1&25.5&\multicolumn{1}{c|}{\textbf{24.5}}&\underline{\textbf{35.3}}&23.6\\
        \multicolumn{1}{c|}{}&\multicolumn{1}{c|}{FGSM ($\epsilon_0=16$)} &
            18.5&\textbf{23.2}&\multicolumn{1}{c|}{19.2}&\underline{\textbf{30.0}}&19.3\\
        \multicolumn{1}{c|}{}&\multicolumn{1}{c|}{FGSM ($\epsilon_0=32$)} &
            11.1&\textbf{18.7}&\multicolumn{1}{c|}{15.7}&\underline{\textbf{24.4}}&16.4\\
        \multicolumn{1}{c|}{}&\multicolumn{1}{c|}{PGD-$l_\infty$} &
        0.0&0.0&\multicolumn{1}{c|}{\underline{\textbf{13.9}}}&7.2&\textbf{9.0}\\
        \multicolumn{1}{c|}{}&\multicolumn{1}{c|}{PGD-$l_2$} &
        0.2&1.0&\multicolumn{1}{c|}{\underline{\textbf{26.9}}}&15.7&\textbf{21.5}\\
        \cline{2-7}
        \multicolumn{1}{c|}{}&\multicolumn{1}{c|}{Corrupted-1} &   
        69.0&\underline{\textbf{72.7}}&\multicolumn{1}{c|}{65.7}&\textbf{69.8}&67.0\\
        \multicolumn{1}{c|}{}&\multicolumn{1}{c|}{Corrupted-2} &   
        65.2&\underline{\textbf{70.7}}&\multicolumn{1}{c|}{64.1}&\textbf{67.3}&65.3\\
        \multicolumn{1}{c|}{}&\multicolumn{1}{c|}{Corrupted-3} &   
        62.2&\underline{\textbf{68.8}}&\multicolumn{1}{c|}{62.7}&\textbf{65.3}&63.7\\
        \multicolumn{1}{c|}{}&\multicolumn{1}{c|}{Corrupted-4} &   
        58.6&\underline{\textbf{66.4}}&\multicolumn{1}{c|}{61.2}&\textbf{62.9}&61.9\\
        \multicolumn{1}{c|}{}&\multicolumn{1}{c|}{Corrupted-5} &   
        53.7&\underline{\textbf{62.6}}&\multicolumn{1}{c|}{58.1}&\textbf{58.7}&58.6\\
        \hline
    \end{tabular}
\end{table}

\setlength{\tabcolsep}{7pt}
\begin{table}[p]
    \centering
    \caption{The classification accuracy (\%) of ResNet-18, WideResNet-40-2, and DenseNet classifiers trained on CIFAR-10 with different combination of images. The PGD attack with $\epsilon=8/255$ was used in training. The top-2 results are indicated in bold while the best results are underlined.}
    \label{table:cifar10_pgd}
    \begin{tabular}{ccccccc}
        \\
        \hline
        & \multicolumn{1}{c|}{}& \multicolumn{5}{c}{Combination of Training Data} \\
        &\multicolumn{1}{c|}{}&Clean&APR&\multicolumn{1}{c|}{C\&Adv}&C\&AA&C\&AP\\
        \hline
        \multicolumn{1}{c|}{\multirow{11}{*}{ResNet-18}}&\multicolumn{1}{c|}{Clean} &   
            \underline{\textbf{94.8}}&\textbf{94.6}&\multicolumn{1}{c|}{85.0}&90.1&88.1\\
        \cline{2-7}
        \multicolumn{1}{c|}{}&\multicolumn{1}{c|}{FGSM ($\epsilon_0=4$)} &
            71.6&70.9&\multicolumn{1}{c|}{67.9}&\underline{\textbf{76.7}}&\textbf{73.9}\\
        \multicolumn{1}{c|}{}&\multicolumn{1}{c|}{FGSM ($\epsilon_0=8$)} &
            65.6&66.2&\multicolumn{1}{c|}{59.9}&\underline{\textbf{72.8}}&\textbf{69.4}\\
        \multicolumn{1}{c|}{}&\multicolumn{1}{c|}{FGSM ($\epsilon_0=16$)} &
            58.1&61.3&\multicolumn{1}{c|}{53.5}&\underline{\textbf{68.2}}&\textbf{63.1}\\
        \multicolumn{1}{c|}{}&\multicolumn{1}{c|}{FGSM ($\epsilon_0=32$)} &
            44.7&54.3&\multicolumn{1}{c|}{47.2}&\underline{\textbf{62.6}}&\textbf{55.9}\\
        \multicolumn{1}{c|}{}&\multicolumn{1}{c|}{PGD-$l_\infty$} &
        0.0&0.3&\multicolumn{1}{c|}{\underline{\textbf{53.9}}}&\textbf{47.9}&44.0\\
        \multicolumn{1}{c|}{}&\multicolumn{1}{c|}{PGD-$l_2$} &
        1.1&7.7&\multicolumn{1}{c|}{57.4}&\underline{\textbf{60.2}}&\textbf{58.9}\\
        \cline{2-7}
        \multicolumn{1}{c|}{}&\multicolumn{1}{c|}{Corrupted-1} &   
            \textbf{91.2}&\underline{\textbf{92.7}}&\multicolumn{1}{c|}{85.6}&89.3&87.9\\
        \multicolumn{1}{c|}{}&\multicolumn{1}{c|}{Corrupted-2} &   
            \textbf{89.0}&\underline{\textbf{91.4}}&\multicolumn{1}{c|}{84.4}&88.3&86.8\\
        \multicolumn{1}{c|}{}&\multicolumn{1}{c|}{Corrupted-3} &   
            87.1&\underline{\textbf{90.3}}&\multicolumn{1}{c|}{83.2}&\textbf{87.2}&85.8\\
        \multicolumn{1}{c|}{}&\multicolumn{1}{c|}{Corrupted-4} &   
            84.5&\underline{\textbf{88.5}}&\multicolumn{1}{c|}{81.8}&\textbf{85.7}&84.3\\
        \multicolumn{1}{c|}{}&\multicolumn{1}{c|}{Corrupted-5} &   
            80.9&\underline{\textbf{85.9}}&\multicolumn{1}{c|}{79.1}&\textbf{83.1}&81.8\\
        \hline 
        \hline
        \multicolumn{1}{c|}{\multirow{11}{*}{WideResNet}}&\multicolumn{1}{c|}{Clean} &   
            \textbf{94.1}&\underline{\textbf{94.3}}&\multicolumn{1}{c|}{83.8}&88.5&86.2\\
        \cline{2-7}
        \multicolumn{1}{c|}{}&\multicolumn{1}{c|}{FGSM ($\epsilon_0=4$)} &
            \underline{\textbf{72.5}}&69.7&\multicolumn{1}{c|}{65.7}&\textbf{70.9}&68.2\\
        \multicolumn{1}{c|}{}&\multicolumn{1}{c|}{FGSM ($\epsilon_0=8$)} &
            \underline{\textbf{66.8}}&64.5&\multicolumn{1}{c|}{55.2}&\textbf{66.2}&62.8\\
        \multicolumn{1}{c|}{}&\multicolumn{1}{c|}{FGSM ($\epsilon_0=16$)} &
            58.8&\textbf{60.0}&\multicolumn{1}{c|}{45.2}&\underline{\textbf{61.4}}&57.5\\
        \multicolumn{1}{c|}{}&\multicolumn{1}{c|}{FGSM ($\epsilon_0=32$)} &
            43.5&41.0&\multicolumn{1}{c|}{38.3}&\underline{\textbf{56.7}}&\textbf{51.9}\\
        \multicolumn{1}{c|}{}&\multicolumn{1}{c|}{PGD-$l_\infty$} &
        0.2&0.3&\multicolumn{1}{c|}{\underline{\textbf{46.2}}}&\textbf{42.6}&39.6\\
        \multicolumn{1}{c|}{}&\multicolumn{1}{c|}{PGD-$l_2$} &
            2.4&7.0&\multicolumn{1}{c|}{\textbf{52.1}}&\underline{\textbf{52.7}}&51.6\\
        \cline{2-7}
        \multicolumn{1}{c|}{}&\multicolumn{1}{c|}{Corrupted-1} &   
            \textbf{90.4}&\underline{\textbf{92.2}}&\multicolumn{1}{c|}{84.5}&88.0&86.4\\
        \multicolumn{1}{c|}{}&\multicolumn{1}{c|}{Corrupted-2} &   
            \textbf{88.1}&\underline{\textbf{90.9}}&\multicolumn{1}{c|}{83.3}&86.9&85.3\\
        \multicolumn{1}{c|}{}&\multicolumn{1}{c|}{Corrupted-3} &   
            \textbf{85.9}&\underline{\textbf{89.7}}&\multicolumn{1}{c|}{82.2}&85.7&84.1\\
        \multicolumn{1}{c|}{}&\multicolumn{1}{c|}{Corrupted-4} &   
            83.2&\underline{\textbf{87.7}}&\multicolumn{1}{c|}{80.7}&\textbf{84.1}&82.5\\
        \multicolumn{1}{c|}{}&\multicolumn{1}{c|}{Corrupted-5} &   
            79.3&\underline{\textbf{85.1}}&\multicolumn{1}{c|}{77.7}&\textbf{81.3}&79.7\\
        \hline 
        \hline
        \multicolumn{1}{c|}{\multirow{11}{*}{DenseNet}}&\multicolumn{1}{c|}{Clean} &   
            \textbf{93.4}&\underline{\textbf{93.7}}&\multicolumn{1}{c|}{83.2}&90.0&87.0\\
        \cline{2-7}
        \multicolumn{1}{c|}{}&\multicolumn{1}{c|}{FGSM ($\epsilon_0=4$)} &
            62.6&\underline{\textbf{64.2}}&\multicolumn{1}{c|}{63.7}&\textbf{65.1}&62.4\\
        \multicolumn{1}{c|}{}&\multicolumn{1}{c|}{FGSM ($\epsilon_0=8$)} &
            \textbf{57.1}&\underline{\textbf{59.2}}&\multicolumn{1}{c|}{49.9}&56.5&53.3\\
        \multicolumn{1}{c|}{}&\multicolumn{1}{c|}{FGSM ($\epsilon_0=16$)} &
            \textbf{50.2}&\underline{\textbf{55.6}}&\multicolumn{1}{c|}{38.3}&50.1&46.1\\
        \multicolumn{1}{c|}{}&\multicolumn{1}{c|}{FGSM ($\epsilon_0=32$)} &
            39.1&\underline{\textbf{49.3}}&\multicolumn{1}{c|}{34.5}&\textbf{45.6}&41.3\\
        \multicolumn{1}{c|}{}&\multicolumn{1}{c|}{PGD-$l_\infty$} &
        0.0&0.0&\multicolumn{1}{c|}{\underline{\textbf{37.3}}}&28.9&\textbf{30.2}\\
        \multicolumn{1}{c|}{}&\multicolumn{1}{c|}{PGD-$l_2$} &
        0.1&1.5&\multicolumn{1}{c|}{\underline{\textbf{49.5}}}&\textbf{47.5}&47.2\\
        \cline{2-7}
        \multicolumn{1}{c|}{}&\multicolumn{1}{c|}{Corrupted-1} &   
        \textbf{89.5}&\underline{\textbf{91.7}}&\multicolumn{1}{c|}{84.1}&89.2&87.1\\
        \multicolumn{1}{c|}{}&\multicolumn{1}{c|}{Corrupted-2} &   
        87.1&\underline{\textbf{90.3}}&\multicolumn{1}{c|}{82.9}&\textbf{88.0}&85.9\\
        \multicolumn{1}{c|}{}&\multicolumn{1}{c|}{Corrupted-3} &   
        85.0&\underline{\textbf{89.0}}&\multicolumn{1}{c|}{81.7}&\textbf{86.9}&84.8\\
        \multicolumn{1}{c|}{}&\multicolumn{1}{c|}{Corrupted-4} &   
        82.2&\underline{\textbf{87.1}}&\multicolumn{1}{c|}{80.4}&\textbf{85.4}&83.3\\
        \multicolumn{1}{c|}{}&\multicolumn{1}{c|}{Corrupted-5} &   
        78.3&\underline{\textbf{84.5}}&\multicolumn{1}{c|}{77.7}&\textbf{82.8}&80.7\\
        \hline
    \end{tabular}
\end{table}

\setlength{\tabcolsep}{7pt}
\begin{table}[p]
    \centering
    \caption{The classification accuracy (\%) of ResNet-18, WideResNet-40-2, and DenseNet classifiers trained on CIFAR-100 with different combination of images. The PGD attack with $\epsilon=8/255$ was used in training. The top-2 results are indicated in bold while the best results are underlined.}
    \label{table:cifar100_pgd}
    \begin{tabular}{ccccccc}
        \\
        \hline
        & \multicolumn{1}{c|}{}& \multicolumn{5}{c}{Combination of Training Data} \\
        &\multicolumn{1}{c|}{}&Clean&APR&\multicolumn{1}{c|}{C\&Adv}&C\&AA&C\&AP\\
        \hline
        \multicolumn{1}{c|}{\multirow{11}{*}{ResNet-18}}&\multicolumn{1}{c|}{Clean} &   
            \underline{\textbf{75.9}}&\textbf{74.5}&\multicolumn{1}{c|}{56.7}&63.8&61.5\\
        \cline{2-7}
        \multicolumn{1}{c|}{}&\multicolumn{1}{c|}{FGSM ($\epsilon_0=4$)} &
            36.6&36.9&\multicolumn{1}{c|}{37.0}&\underline{\textbf{41.9}}&\textbf{39.6}\\
        \multicolumn{1}{c|}{}&\multicolumn{1}{c|}{FGSM ($\epsilon_0=8$)} &
            30.3&32.7&\multicolumn{1}{c|}{29.7}&\underline{\textbf{36.5}}&\textbf{34.6}\\
        \multicolumn{1}{c|}{}&\multicolumn{1}{c|}{FGSM ($\epsilon_0=16$)} &
            24.0&29.2&\multicolumn{1}{c|}{24.6}&\underline{\textbf{31.8}}&\textbf{30.0}\\
        \multicolumn{1}{c|}{}&\multicolumn{1}{c|}{FGSM ($\epsilon_0=32$)} &
            16.4&23.1&\multicolumn{1}{c|}{20.3}&\underline{\textbf{26.5}}&\textbf{24.4}\\
        \multicolumn{1}{c|}{}&\multicolumn{1}{c|}{PGD-$l_\infty$} &
        0.0&0.0&\multicolumn{1}{c|}{\underline{\textbf{20.8}}}&\textbf{20.1}&19.4\\
        \multicolumn{1}{c|}{}&\multicolumn{1}{c|}{PGD-$l_2$} &
        0.2&0.4&\multicolumn{1}{c|}{28.3}&\underline{\textbf{30.6}}&\textbf{29.9}\\
        \cline{2-7}
        \multicolumn{1}{c|}{}&\multicolumn{1}{c|}{Corrupted-1} &   
            \textbf{74.3}&\underline{\textbf{75.9}}&\multicolumn{1}{c|}{62.9}&68.4&66.7\\
        \multicolumn{1}{c|}{}&\multicolumn{1}{c|}{Corrupted-2} &   
            \textbf{71.3}&\underline{\textbf{74.2}}&\multicolumn{1}{c|}{61.6}&67.1&65.3\\
        \multicolumn{1}{c|}{}&\multicolumn{1}{c|}{Corrupted-3} &   
            \textbf{68.7}&\underline{\textbf{72.6}}&\multicolumn{1}{c|}{60.3}&65.7&63.9\\
        \multicolumn{1}{c|}{}&\multicolumn{1}{c|}{Corrupted-4} &   
            \textbf{65.6}&\underline{\textbf{70.2}}&\multicolumn{1}{c|}{58.8}&63.9&62.3\\
        \multicolumn{1}{c|}{}&\multicolumn{1}{c|}{Corrupted-5} &   
            \textbf{61.0}&\underline{\textbf{66.9}}&\multicolumn{1}{c|}{55.9}&\textbf{61.0}&59.5\\
        \hline 
        \hline
        \multicolumn{1}{c|}{\multirow{11}{*}{WideResNet}}&\multicolumn{1}{c|}{Clean} &   
            \underline{\textbf{72.5}}&\textbf{72.0}&\multicolumn{1}{c|}{57.1}&60.7&59.4\\
        \hline
        \multicolumn{1}{c|}{}&\multicolumn{1}{c|}{FGSM ($\epsilon_0=4$)} &
            32.8&35.5&\multicolumn{1}{c|}{\underline{\textbf{36.2}}}&\textbf{36.1}&35.0\\
        \multicolumn{1}{c|}{}&\multicolumn{1}{c|}{FGSM ($\epsilon_0=8$)} &
            27.5&\underline{\textbf{31.7}}&\multicolumn{1}{c|}{27.0}&\textbf{30.5}&29.1\\
        \multicolumn{1}{c|}{}&\multicolumn{1}{c|}{FGSM ($\epsilon_0=16$)} &
            20.5&\underline{\textbf{27.2}}&\multicolumn{1}{c|}{20.5}&\textbf{26.3}&24.4\\
        \multicolumn{1}{c|}{}&\multicolumn{1}{c|}{FGSM ($\epsilon_0=32$)} &
            12.7&\textbf{20.4}&\multicolumn{1}{c|}{16.7}&\underline{\textbf{22.4}}&20.2\\
        \multicolumn{1}{c|}{}&\multicolumn{1}{c|}{PGD-$l_\infty$} &
            0.0&0.0&\multicolumn{1}{c|}{\underline{\textbf{18.7}}}&14.8&\textbf{15.5}\\
        \multicolumn{1}{c|}{}&\multicolumn{1}{c|}{PGD-$l_2$} &
            0.1&1.0&\multicolumn{1}{c|}{\underline{\textbf{26.5}}}&22.8&\textbf{23.9}\\
        \cline{2-7}
        \multicolumn{1}{c|}{}&\multicolumn{1}{c|}{Corrupted-1} &   
        \textbf{70.3}&\underline{\textbf{73.5}}&\multicolumn{1}{c|}{55.5}&65.7&64.9\\
        \multicolumn{1}{c|}{}&\multicolumn{1}{c|}{Corrupted-2} &   
        \textbf{66.8}&\underline{\textbf{71.7}}&\multicolumn{1}{c|}{54.7}&64.2&63.3\\
        \multicolumn{1}{c|}{}&\multicolumn{1}{c|}{Corrupted-3} &   
        \textbf{63.9}&\underline{\textbf{69.8}}&\multicolumn{1}{c|}{53.3}&62.8&61.8\\
        \multicolumn{1}{c|}{}&\multicolumn{1}{c|}{Corrupted-4} &   
        60.3&\underline{\textbf{67.2}}&\multicolumn{1}{c|}{51.7}&\textbf{60.9}&59.9\\
        \multicolumn{1}{c|}{}&\multicolumn{1}{c|}{Corrupted-5} &   
        55.2&\underline{\textbf{63.4}}&\multicolumn{1}{c|}{48.8}&\textbf{57.7}&56.9\\
        \hline
        \hline
        \multicolumn{1}{c|}{\multirow{11}{*}{DenseNet}}&\multicolumn{1}{c|}{Clean} &   
            \textbf{71.3}&\underline{\textbf{71.5}}&\multicolumn{1}{c|}{56.8}&67.9&62.2\\
        \cline{2-7}
        \multicolumn{1}{c|}{}&\multicolumn{1}{c|}{FGSM ($\epsilon_0=4$)} &
            28.3&27.7&\multicolumn{1}{c|}{\textbf{36.0}}&\underline{\textbf{43.0}}&32.0\\
        \multicolumn{1}{c|}{}&\multicolumn{1}{c|}{FGSM ($\epsilon_0=8$)} &
            24.1&\textbf{25.5}&\multicolumn{1}{c|}{24.9}&\underline{\textbf{36.0}}&23.4\\
        \multicolumn{1}{c|}{}&\multicolumn{1}{c|}{FGSM ($\epsilon_0=16$)} &
            18.5&\textbf{23.2}&\multicolumn{1}{c|}{17.1}&\underline{\textbf{28.8}}&18.7\\
        \multicolumn{1}{c|}{}&\multicolumn{1}{c|}{FGSM ($\epsilon_0=32$)} &
            11.1&\textbf{18.7}&\multicolumn{1}{c|}{12.7}&\underline{\textbf{23.4}}&15.3\\
        \multicolumn{1}{c|}{}&\multicolumn{1}{c|}{PGD-$l_\infty$} &
        0.0&0.0&\multicolumn{1}{c|}{\underline{\textbf{17.4}}}&\textbf{15.1}&11.6\\
        \multicolumn{1}{c|}{}&\multicolumn{1}{c|}{PGD-$l_2$} &
        0.2&1.0&\multicolumn{1}{c|}{\underline{\textbf{29.7}}}&21.4&\textbf{24.6}\\
        \cline{2-7}
        \multicolumn{1}{c|}{}&\multicolumn{1}{c|}{Corrupted-1} &   
        69.0&\underline{\textbf{72.7}}&\multicolumn{1}{c|}{63.1}&\textbf{70.6}&66.9\\
        \multicolumn{1}{c|}{}&\multicolumn{1}{c|}{Corrupted-2} &   
        65.2&\underline{\textbf{70.7}}&\multicolumn{1}{c|}{61.7}&\textbf{68.8}&65.3\\
        \multicolumn{1}{c|}{}&\multicolumn{1}{c|}{Corrupted-3} &   
        62.2&\underline{\textbf{68.8}}&\multicolumn{1}{c|}{60.4}&\textbf{67.2}&63.7\\
        \multicolumn{1}{c|}{}&\multicolumn{1}{c|}{Corrupted-4} &   
        58.6&\underline{\textbf{66.4}}&\multicolumn{1}{c|}{58.8}&\textbf{65.2}&62.0\\
        \multicolumn{1}{c|}{}&\multicolumn{1}{c|}{Corrupted-5} &   
        53.7&\underline{\textbf{62.6}}&\multicolumn{1}{c|}{56.0}&\textbf{61.7}&58.7\\
        \hline
    \end{tabular}
\end{table}

\setlength{\tabcolsep}{7pt}
\begin{table}[t]
    \centering
    \caption{The classification accuracy (\%) of ResNet-18 classifiers trained on Tiny ImageNet with different combination of images. The FGSM attack with $\epsilon=8/255$ was used in training. The top-2 results are indicated in bold while the best results are underlined.}
    \label{table:imagenet}
    \begin{tabular}{ccccccc}
        \\
        \hline
        & \multicolumn{1}{c|}{}& \multicolumn{5}{c}{Combination of Training Data} \\
        &\multicolumn{1}{c|}{}&Clean&APR&\multicolumn{1}{c|}{C\&Adv}&C\&AA&C\&AP\\
        \hline
        \multicolumn{1}{c|}{\multirow{11}{*}{ResNet-18}}&\multicolumn{1}{c|}{Clean} &   
            \textbf{46.2}&\underline{\textbf{46.5}}&\multicolumn{1}{c|}{42.7}&\textbf{46.2}&44.5\\
        \hline
        \multicolumn{1}{c|}{}&\multicolumn{1}{c|}{FGSM ($\epsilon_0=4$)} &
            0.8&0.8&\multicolumn{1}{c|}{\underline{\textbf{17.1}}}&12.9&\textbf{13.1}\\
        \multicolumn{1}{c|}{}&\multicolumn{1}{c|}{FGSM ($\epsilon_0=8$)} &
            0.4&0.5&\multicolumn{1}{c|}{\underline{\textbf{8.2}}}&\textbf{4.9}&4.5\\
        \multicolumn{1}{c|}{}&\multicolumn{1}{c|}{FGSM ($\epsilon_0=16$)} &
            0.6&0.6&\multicolumn{1}{c|}{\underline{\textbf{2.8}}}&\textbf{1.4}&1.0\\
        \multicolumn{1}{c|}{}&\multicolumn{1}{c|}{FGSM ($\epsilon_0=32$)} &
            \textbf{0.7}&\textbf{0.7}&\multicolumn{1}{c|}{\underline{\textbf{0.8}}}&0.6&0.5\\
        \multicolumn{1}{c|}{}&\multicolumn{1}{c|}{PGD-$l_\infty$} &
            0.0&0.0&\multicolumn{1}{c|}{\underline{\textbf{1.9}}}&0.5&\textbf{0.6}\\
        \multicolumn{1}{c|}{}&\multicolumn{1}{c|}{PGD-$l_2$} &
            0.1&0.2&\multicolumn{1}{c|}{\underline{\textbf{26.4}}}&\textbf{23.3}&23.0\\
        \cline{2-7}
        \multicolumn{1}{c|}{}&\multicolumn{1}{c|}{Corrupted-1} &   
        \textbf{40.7}&\underline{\textbf{41.1}}&\multicolumn{1}{c|}{38.0}&40.3&38.0\\
        \multicolumn{1}{c|}{}&\multicolumn{1}{c|}{Corrupted-2} &   
        \textbf{37.8}&\underline{\textbf{38.1}}&\multicolumn{1}{c|}{35.8}&37.7&35.7\\
        \multicolumn{1}{c|}{}&\multicolumn{1}{c|}{Corrupted-3} &   
        34.1&\textbf{34.3}&\multicolumn{1}{c|}{33.0}&\underline{\textbf{34.4}}&32.9\\
        \multicolumn{1}{c|}{}&\multicolumn{1}{c|}{Corrupted-4} &   
        \textbf{30.6}&\textbf{30.6}&\multicolumn{1}{c|}{30.2}&\underline{\textbf{31.2}}&30.1\\
        \multicolumn{1}{c|}{}&\multicolumn{1}{c|}{Corrupted-5} &   
        27.3&\textbf{27.4}&\multicolumn{1}{c|}{27.2}&\underline{\textbf{28.0}}&27.1\\
        \hline
    \end{tabular}
\end{table}

\section{Comparison with adversarial training with APR}
In the main text, the APR data augmentation was applied to clean images, aimed to encourage CNN classifiers to focus on the phase spectrum of images and hence improve the robustness against common corruptions. 
\citet{chen2021apr} also showed that APR can be applied to FGSM adversarial training to improve adversarial robustness.
The crucial difference from our method is that APR was applied to clean images prior to the FGSM perturbations, while our method swaps the amplitude spectrum of clean images with that of the adversarial images. 
When these APR images are adversarially perturbed, the classifiers are encouraged to learn from these adversarially perturbed phase spectra and hence deliver better robustness against adversarial perturbations. 
In contrast, the proposed method, adversarial amplitude swap, is used to recombine the frequency spectra of clean and adversarial images to generate adversarial amplitude and adversarial phase images. These images contain only one type of frequency spectrum of adversarial images and the other type of frequency spectrum of clean images. With the adversarial amplitude images, classifiers are encouraged to learn from the phase spectrum of clean images and the amplitude spectrum of adversarial images, resulting in general robustness against both common corruptions and adversarial perturbations. 

To demonstrate that this difference leads to a substantial gap in the general robustness, we conducted experiments on CIFAR-10 and CIFAR-100 with the same training setup in Section~\ref{sec:cifar}. FGSM adversarial training was used to train the baseline model, in which the FGSM perturbed images were used as training images. 
Table~\ref{table:comparison_apr} shows the classification accuracy of WideResnet models trained on CIFAR-10 and CIFAR-100 datasets. Note that the results of the baseline model trained with adverarial~(Adv) images, the model trained with adversarial amplitude~(AA) images, and the model trained with adversarial phase~(AP) images on the CIFAR-10 dataset were also shown in Table~\ablation{}. 
Notably, the baseline model overfitted FGSM perturbations, while others did not. 
When APR was applied on adversarial images~APR improved model robustness against adversarial perturbations but failed to do so on the robustness against common corruptions. 
Adversarial phase images delivered similar performance as the Adv-APR but the clean accuracy was 3.7\% higher. 
In contrast, adversarial amplitude images improved both the robustness against common corruptions and adversarial perturbations compared to the baseline model. These results further showed that adversarial amplitude images can serve as a better data augmentation method to achieve general robustness against both common corruptions and adversarial perturbations even in an adversarial training setup. 

\setlength{\tabcolsep}{7pt}
\begin{table}[t]
    \centering
    \caption{The classification accuracy (\%) of WideResNet-40-2 classifiers trained on CIFAR-10 and CIFAR-100 with adversarial examples and several data augmentation methods. The FGSM attack with $\epsilon=8/255$ was used in training. The top-2 results are indicated in bold while the best results are underlined.}
    \label{table:comparison_apr}
    \begin{tabular}{cccccc}
        \\
        \hline
        & \multicolumn{1}{c|}{}& \multicolumn{4}{c}{Training Data} \\
        \multicolumn{1}{c|}{}&\multicolumn{1}{c|}{}&Adv&\multicolumn{1}{c|}{Adv-APR}&AA&AP\\
        \hline
        \multicolumn{1}{c|}{\multirow{11}{*}{CIFAR-10}}&\multicolumn{1}{c|}{Clean} &   
            86.8&\multicolumn{1}{c|}{83.2}&\underline{\textbf{89.0}}&\textbf{86.9}\\
        \hline
        \multicolumn{1}{c|}{}&\multicolumn{1}{c|}{FGSM ($\epsilon_0=4$)} &
            \underline{\textbf{88.7}}&\multicolumn{1}{c|}{63.7}&\textbf{74.7}&71.2\\
        \multicolumn{1}{c|}{}&\multicolumn{1}{c|}{FGSM ($\epsilon_0=8$)} &
            \underline{\textbf{88.6}}&\multicolumn{1}{c|}{54.7}&\textbf{70.3}&66.8\\
        \multicolumn{1}{c|}{}&\multicolumn{1}{c|}{FGSM ($\epsilon_0=16$)} &
            \underline{\textbf{87.6}}&\multicolumn{1}{c|}{47.7}&\textbf{65.3}&61.6\\
        \multicolumn{1}{c|}{}&\multicolumn{1}{c|}{FGSM ($\epsilon_0=32$)} &
            \underline{\textbf{85.3}}&\multicolumn{1}{c|}{43.9}&\textbf{59.8}&54.9\\
        \multicolumn{1}{c|}{}&\multicolumn{1}{c|}{PGD-$l_\infty$} &
            19.1&\multicolumn{1}{c|}{36.4}&\textbf{40.4}&\underline{\textbf{40.8}}\\
        \multicolumn{1}{c|}{}&\multicolumn{1}{c|}{PGD-$l_2$} &
            12.3&\multicolumn{1}{c|}{\textbf{46.8}}&\underline{\textbf{53.5}}&\underline{\textbf{53.6}}\\
        \cline{2-6}
        \multicolumn{1}{c|}{}&\multicolumn{1}{c|}{Corrupted-1} &   
        86.5&\multicolumn{1}{c|}{84.1}&\underline{\textbf{88.4}}&\textbf{86.7}\\
        \multicolumn{1}{c|}{}&\multicolumn{1}{c|}{Corrupted-2} &   
        85.1&\multicolumn{1}{c|}{83.0}&\underline{\textbf{87.2}}&\textbf{85.5}\\
        \multicolumn{1}{c|}{}&\multicolumn{1}{c|}{Corrupted-3} &   
        83.5&\multicolumn{1}{c|}{82.1}&\underline{\textbf{86.1}}&\textbf{84.4}\\
        \multicolumn{1}{c|}{}&\multicolumn{1}{c|}{Corrupted-4} &   
        81.6&\multicolumn{1}{c|}{80.8}&\underline{\textbf{84.4}}&\textbf{82.8}\\
        \multicolumn{1}{c|}{}&\multicolumn{1}{c|}{Corrupted-5} &   
        78.0&\multicolumn{1}{c|}{78.2}&\underline{\textbf{81.5}}&\textbf{79.8}\\
        \hline
        \hline
        \multicolumn{1}{c|}{\multirow{11}{*}{CIFAR-100}}&\multicolumn{1}{c|}{Clean} &   
            \underline{\textbf{64.3}}&\multicolumn{1}{c|}{55.7}&\textbf{61.1}&58.5\\
        \hline
        \multicolumn{1}{c|}{}&\multicolumn{1}{c|}{FGSM ($\epsilon_0=4$)} &
            \underline{\textbf{82.2}}&\multicolumn{1}{c|}{35.3}&\textbf{37.2}&36.1\\
        \multicolumn{1}{c|}{}&\multicolumn{1}{c|}{FGSM ($\epsilon_0=8$)} &
            \underline{\textbf{83.2}}&\multicolumn{1}{c|}{27.1}&\textbf{32.1}&31.1\\
        \multicolumn{1}{c|}{}&\multicolumn{1}{c|}{FGSM ($\epsilon_0=16$)} &
            \underline{\textbf{74.9}}&\multicolumn{1}{c|}{21.9}&\textbf{27.9}&26.1\\
        \multicolumn{1}{c|}{}&\multicolumn{1}{c|}{FGSM ($\epsilon_0=32$)} &
            \underline{\textbf{55.1}}&\multicolumn{1}{c|}{18.7}&\textbf{23.6}&21.3\\
        \multicolumn{1}{c|}{}&\multicolumn{1}{c|}{PGD-$l_\infty$} &
            0.7&\multicolumn{1}{c|}{\underline{\textbf{16.8}}}&14.3&\textbf{15.9}\\
        \multicolumn{1}{c|}{}&\multicolumn{1}{c|}{PGD-$l_2$} &
            0.4&\multicolumn{1}{c|}{\underline{\textbf{24.7}}}&22.7&\textbf{24.6}\\
        \cline{2-6}
        \multicolumn{1}{c|}{}&\multicolumn{1}{c|}{Corrupted-1} &   
        \underline{\textbf{67.4}}&\multicolumn{1}{c|}{62.2}&\textbf{65.7}&64.1\\
        \multicolumn{1}{c|}{}&\multicolumn{1}{c|}{Corrupted-2} &   
        \underline{\textbf{65.1}}&\multicolumn{1}{c|}{60.9}&\textbf{64.0}&62.7\\
        \multicolumn{1}{c|}{}&\multicolumn{1}{c|}{Corrupted-3} &   
        \underline{\textbf{63.1}}&\multicolumn{1}{c|}{59.8}&\textbf{62.4}&61.2\\
        \multicolumn{1}{c|}{}&\multicolumn{1}{c|}{Corrupted-4} &   
        \textbf{60.4}&\multicolumn{1}{c|}{58.4}&\underline{\textbf{60.6}}&59.6\\
        \multicolumn{1}{c|}{}&\multicolumn{1}{c|}{Corrupted-5} &   
        56.2&\multicolumn{1}{c|}{55.7}&\underline{\textbf{57.4}}&\textbf{56.4}\\
        \hline
    \end{tabular}
\end{table}
